\documentclass{article} 
\usepackage{iclr2027_conference,times}

\usepackage{amsmath,amsfonts,bm}

\def\eqref#1{equation~\ref{#1}}

\def\1{\bm{1}}

\DeclareMathAlphabet{\mathsfit}{\encodingdefault}{\sfdefault}{m}{sl}
\SetMathAlphabet{\mathsfit}{bold}{\encodingdefault}{\sfdefault}{bx}{n}

\usepackage{graphicx}
\usepackage{booktabs}
\usepackage{multirow}
\usepackage{textcomp}
\usepackage{xcolor}
\usepackage{tcolorbox}
\tcbuselibrary{skins,breakable}
\usepackage{amssymb}
\usepackage{paralist}
\usepackage{hyperref}
\usepackage{url}

\title{LISA: Linear-Indexed Sparse Attention \\ for Efficient Long Chain-of-Thought Reasoning}

\author{
  \textbf{Yu Zhao\textsuperscript{1}},
  \textbf{Zekun Zhang\textsuperscript{2}},
  \textbf{Fan Jiang\textsuperscript{3}},
  \textbf{Bo Zeng\textsuperscript{1}},
  \textbf{Xintong Wang\textsuperscript{1}},
  \textbf{Chonghan Liu\textsuperscript{4}} \\
  \textbf{Shimin Shan\textsuperscript{2}},
  \textbf{Yichao Du\textsuperscript{4}\thanks{Corresponding authors.}},
  \textbf{Yu Liu\textsuperscript{2}\footnotemark[1]} \\
  \textsuperscript{1}Alibaba Group \\
  \textsuperscript{2}School of Software Technology, Dalian University of Technology, Dalian, China \\
  \textsuperscript{3}Amazon \\
  \textsuperscript{4}Independent Researcher
}

\iclrfinalcopy 
\begin{document}
\maketitle
\begin{abstract}
Long chain-of-thought (CoT) reasoning models such as DeepSeek-R1 keep lengthening their inference contexts under test-time scaling, yet the $O(n^2)$ complexity of standard self-attention makes long-CoT reasoning costly to deploy in production.
We therefore propose \textbf{LISA} (\textbf{L}inear-\textbf{I}ndexed \textbf{S}parse \textbf{A}ttention), a plug-and-play attention replacement module that requires no pretraining from scratch. LISA integrates two lightweight components in parallel inside the original model --- a Linear Attention branch giving long-range memory at $O(n)$, and a Lightning Indexer that selects the top-$M$ tokens from the full context for a Sparse Self-Attention branch --- and fuses them by gating,
replacing dense attention in each LISA layer with attention over a fixed budget of $M$ selected tokens ($M \ll n$). Because full-attention layers are retained periodically and the Indexer still scores the full context, the overall sequence-level complexity class remains quadratic; the practical gain is a measured reduction in wall-clock latency rather than a change of asymptotic class.
In a two-stage pipeline, Stage~1 trains the linear-attention branch with a next-token cross-entropy objective while retaining sliding-window attention and freezing the original model parameters.
Stage~2 replaces that static window with the Indexer, enabling dynamic token selection over broader contexts; it is trained by a novel per-head KL divergence loss that aligns its selection behavior with the teacher's attention patterns.
On DeepSeek-distilled-Qwen models, LISA cuts measured eager-mode mean sample latency by 23.9\% ($1.31\times$) on the mathematical benchmarks for the 1.5B model, while improving equal-weight average accuracy by 4.2~percentage points over the full-attention baseline across nine benchmarks. A secondary decoding-FLOPs estimate (Appendix~\ref{app:flops}), counting Indexer full-context scoring, is consistent with a moderate constant-factor saving at long context (about 26\% for 1.5B at 16K).
\end{abstract}
\section{Introduction}

Chain-of-Thought (CoT)~\citep{wei2022chain,kojima2022large} has become a core technique for enhancing the reasoning ability of large language models (LLMs) on complex tasks. By prompting step-by-step reasoning, CoT enables LLMs to decompose complex problems into simpler subtasks, thus improving their problem-solving capabilities~\citep{yao2023tree,wang2023selfconsistency,zhou2023least}. Recent studies, including OpenAI o1~\citep{openai2026openaio1card}, DeepSeek-R1~\citep{deepseek2025r1}, and other reasoning models~\citep{qwen2024qwq,zhao2024marcoo1openreasoningmodels}, demonstrate that scaling up CoT length can further enhance the reasoning abilities of LLMs.

However, since most current LLMs are built upon the Transformer architecture~\citep{vaswani2017attention}, the computational complexity of their attention grows quadratically with context length, and the memory overhead of their KV-cache increases linearly. Hence, generating long CoT substantially increases the computational and memory cost of LLMs, limiting their practical efficiency on complex reasoning tasks.

To improve reasoning efficiency, prior work shortens the chain itself via prompting~\citep{han2024tokenbudget,ma2025reasoning}, SFT~\citep{liu2024skip,munkhbat2025selftraining} or RL~\citep{aggarwal2025l1,shen2025dast}, but this often costs accuracy, since shortening conflicts with test-time scaling~\citep{jin2024impact,merrill2024expressive,openai2026openaio1card}. Others re-express the same reasoning more concisely~\citep{ma2025cotvalve,kang2025c3ot,xia2025tokenskip}, risking the loss of decisive steps~\citep{wang2025r1compress}. \citet{yan2025inftythink} instead interleave short reasoning segments with intermediate summaries to keep context bounded, but require reconstructing training data into this iterative format. None of these changes the quadratic attention that makes long chains expensive in the first place.

In this paper, we propose an efficient hybrid inference framework \textbf{LISA} (\textbf{L}inear-\textbf{I}ndexed \textbf{S}parse \textbf{A}ttention). We replace the standard self-attention of a pretrained model with a hybrid attention architecture: a global linear attention vector maintains long-range context state, while a fixed-size self-attention module preserves precise contextual retrieval capability. This design restricts sparse self-attention in each LISA layer to a fixed budget of $M$ selected tokens while retaining precise access to critical context. Our framework requires no complex data preprocessing, does not alter the original model parameters or output format, and achieves plug-and-play improvement in both inference efficiency and quality.

Our key contributions are as follows:
\begin{compactenum}
    \item We propose a \textbf{plug-and-play hybrid attention module (LISA)} that combines global linear attention with fixed-size self-attention, restricting per-step sparse self-attention to a fixed budget of $M$ tokens while preserving precise contextual retrieval.
    \item We design a \textbf{two-stage training strategy}: the first stage cold-starts linear attention via pure cross-entropy loss to avoid overfitting to the teacher distribution; the second stage introduces an indexer module with top-$M$ dynamic selection, augmented with KL loss to ensure stable important-token selection.
  \item We conduct \textbf{extensive experiments} on five mathematical reasoning benchmarks (GSM8K, MATH-500, AMC23, AIME24, AIME25) and four non-mathematical benchmarks (GPQA-D~\citep{rein2024gpqa}, HumanEval~\citep{chen2021humaneval}, LiveCodeBench~\citep{jain2024livecodebenchholisticcontaminationfree} and BigCodeBench~\citep{zhuo2025bigcodebenchbenchmarkingcodegeneration}), demonstrating that LISA cuts measured eager-mode mean sample latency by 23.9\% ($1.31\times$) on the mathematical benchmarks for the 1.5B model and improves its equal-weight average accuracy by 4.2~percentage points over the full self-attention baseline across these nine benchmarks. A secondary decoding-FLOPs estimate that includes Indexer scoring (Appendix~\ref{app:flops}) indicates a moderate constant-factor saving at long context.
\end{compactenum}
\section{Related Work}

\paragraph{Efficient Reasoning Methods.}
A growing body of work seeks to improve the efficiency of CoT reasoning. One line of research uses prompt engineering, SFT, or RL to encourage models to generate shorter CoT sequences~\citep{nayab2024concise,aggarwal2025l1,munkhbat2025selftraining}. However, these methods often compromise reasoning capability, as shortening CoT conflicts with test-time scaling principles. Alternative approaches aim to express CoT more compactly without losing information. LightThinker~\citep{zhang2025lightthinker} proposes a ``thinking step-by-step compression'' paradigm that inserts learnable special tokens into hidden state sequences and compresses the KV states of completed reasoning steps before each generation step. INFTYTHINK~\citep{yan2025inftythink} compresses the processed lengthy context and prior reasoning trajectories into compact summary representations at each step, freeing up context window capacity for subsequent reasoning. Additionally, TokenSkip~\citep{xia2025tokenskip} achieves controllable CoT compression through a learnable token-skipping mechanism, and R1-Compress~\citep{wang2025r1compress} explores long CoT compression via chunk-level compression and search-based optimization. \citet{zhang2026statetransitionframeworkefficientllm} maintain a global state matrix via linear attention to record historical information, enabling the model to retrieve relevant information without explicitly attending to prior reasoning steps. Our work differs from these methods in that we do not compress the output CoT itself but rather accelerate the internal inference process while preserving the full reasoning capability.

\paragraph{KV-Cache Reduction Methods.}
Reducing KV-cache size is another major direction for efficient LLM inference. H2O~\citep{zhang2023h2o} proposes a layer-wise eviction strategy that retains only KV vectors of Heavy-Hitter tokens and recently generated tokens, maintaining a constant-size KV-cache while preserving generation quality. StreamingLLM~\citep{xiao2023streamingllm} discovers the ``attention sink'' phenomenon and proposes retaining only initial and recent tokens' KV vectors to enable efficient streaming generation. SnapKV~\citep{li2024snapkvllmknowslooking} observes that the model can identify important KV pairs before generation and proposes a query-driven KV cache compression method. PyramidKV~\citep{cai2025pyramidkvdynamickvcache} further reveals a pyramidal information funneling structure across layers and proposes layer-wise adaptive KV cache budgets. SepLLM~\citep{chen2025sepllm} compresses each segment into a single separator token and stores only the KV vectors of initial, neighboring, and separator tokens. These methods focus on reducing the stored KV-cache footprint during decoding; our work instead keeps the full KV cache and targets reducing decoding compute and latency for long-reasoning tasks.

\paragraph{Linear Attention.}
To address the quadratic computational cost of softmax attention, many linear attention methods have been proposed to improve training and inference efficiency. Early studies primarily investigated kernel functions for linear attention, such as random feature attention~\citep{peng2021random} and RNN-izing Transformers~\citep{kasai2021finetuning}. Subsequently, several works introduced gating mechanisms to more precisely control information flow in the model state: GLA~\citep{yang2024gated} achieves hardware-efficient training with gated linear attention, HGRN2~\citep{qin2024hgrn2} introduces state-expanded gated linear RNNs, and Liger~\citep{lan2025liger} linearizes large language models into gated recurrent structures. Recent studies have provided theoretical evidence for the Test-Time Training (TTT) property of linear attention~\citep{yang2024delta,sun2025learninglearntesttime,liu2025longhorn,chen2025donotthink}, offering theoretical foundations for designing online learning objectives to enhance sequence modeling capability.
\section{Method}
\label{sec:method}

\subsection{Preliminaries}

\paragraph{Softmax Attention.}
Decoder-only LLMs rely on multi-head softmax attention~\citep{vaswani2017attention}, whose output is a softmax-weighted sum over values. With a KV cache, attending for one newly generated token costs $O(t)$ in the current context length $t$ (ignoring model width); generating a full sequence of length $n$ therefore accumulates to $O(n^2)$. The KV cache itself grows linearly with sequence length, a significant memory bottleneck in long-context settings.

\paragraph{Linear Attention.}
Linear attention~\citep{katharopoulos2020transformers} replaces the exponential kernel with a decomposable feature map $\phi(\cdot)$ and uses the associativity of matrix multiplication to aggregate KV interactions into a single state matrix. While many kernels have been explored~\citep{kasai2021finetuning,peng2021random}, recent studies~\citep{sun2023retentive,yang2024gated} set $\phi(\cdot)$ to the identity and drop the normalization term: $\mathbf{o}_t = \mathbf{q}_t \mathbf{S}_t$, with state $\mathbf{S}_t = \sum_i \mathbf{k}_i^\top \mathbf{v}_i \in \mathbb{R}^{d_k \times d_v}$ accumulating history. This recurrent form gives $O(1)$ per-step computation and constant memory.

\paragraph{Test-Time Training (TTT).}
A line of work~\citep{yang2024delta,sun2025learninglearntesttime,liu2025longhorn,chen2025donotthink} reads $\mathbf{S}_t$ as a fast-adapting parameter updated at every token by lightweight gradient descent. Under the same left-multiply convention as above, a natural objective is $\mathcal{L}(\mathbf{S}) = -\langle \mathbf{k}_t \mathbf{S},\, \mathbf{v}_t \rangle$, with update $\mathbf{S}_t = \mathbf{S}_{t-1} + \beta \mathbf{k}_t^\top \mathbf{v}_t$. At $\beta = 1$ this coincides with the linear-attention state update, i.e.\ training $\mathbf{S}_t$ on that objective. No direct empirical evidence yet shows that linear attention improves reasoning, but its theoretical properties suggest considerable potential.

\subsection{LISA: Linear-Indexed Sparse Attention}

We design LISA to replace the original attention module. Specifically, our method combines a linear attention stream with a fixed-size self-attention stream, where the former records long-range contextual information and the latter preserves precise detail retrieval. A gating mechanism controls the information flow ratio between the two streams. Following related works on linear attention~\citep{li2025minimax}, we replace the standard attention in the teacher model with our linear attention variant while retaining full attention at Transformer blocks $0,7,14,21,27$ (0-indexed)---five of the 28 blocks, with six LISA blocks between consecutive full-attention blocks except before the final full block---to preserve exact retrieval capability~\citep{arora2024simple}. The sparse self-attention after selection costs $O(M)$ per decoded token, but Indexer scoring still scans the full context ($O(t)$ per step with $K$ heads), and the retained dense layers remain $O(t)$; thus LISA reduces attention compute via a smaller selected-token budget while retaining quadratic sequence-level complexity, consistent with the Lightning Indexer analysis in DeepSeek-V3.2~\citep{deepseek2025v32}.

Specifically, we divide the training into two stages.

\begin{figure}[t]
    \centering
    \setlength{\abovecaptionskip}{-0.5cm}
    \setlength{\belowcaptionskip}{-0.3cm}
    \includegraphics[width=0.9\textwidth]{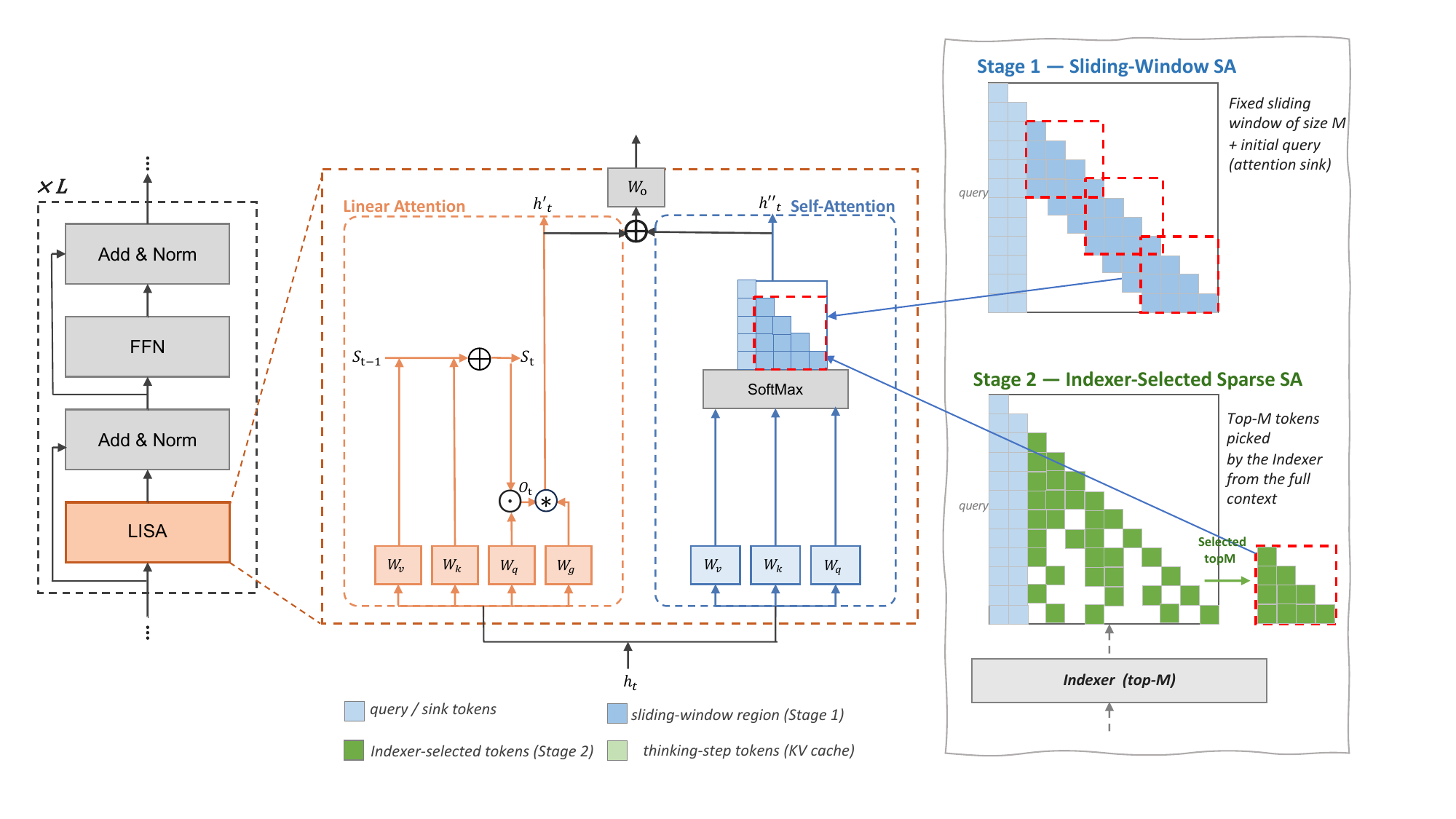}
    \caption{Overview of the LISA architecture.}
    \label{fig:lisa}
\end{figure}

\paragraph{Stage 1: Linear Attention Cold-Start.}
In this stage, we convert the standard self-attention to a fixed-budget sliding window of size $M$, while introducing a linear attention as in Figure~\ref{fig:lisa}.

During training, we freeze all original model parameters and only train the newly added linear attention parameters. We construct an attention block that fuses linear attention and self-attention through a learnable gating parameter:
\begin{equation}
\mathbf{o}_{\mathrm{LISA}} = g \cdot \mathbf{o}_{\mathrm{LA}} + (1 - g) \cdot \mathbf{o}_{\mathrm{SA\_M}},
\end{equation}
where $g$ is the learnable gate, $\mathbf{o}_{\mathrm{LA}}$ is the linear attention output, and $\mathbf{o}_{\mathrm{SA\_M}}$ is the sliding-window self-attention output with window size $M$.

We aim for the LISA output to remain close to the teacher's generation behavior. Rather than using an explicit distillation loss (e.g., KL over attention maps) in Stage~1, we train with standard next-token cross-entropy. Empirically, applying KL over attention distributions in early training led to severe overfitting and degraded downstream generation; we therefore defer attention-distribution matching to the Indexer objective in Stage~2.

\paragraph{Stage 2: Indexer-Guided Token Selection.}
In this stage, we further introduce a lightweight token selector called the \emph{Indexer}. Its design is inspired by the Lightning Indexer in DeepSeek V3.2~\citep{deepseek2025v32} and GLM-5~\citep{zai2026glm5}: it contains only query/key projections and a ReLU dot product, used to select important tokens based on the current context.

Specifically, we maintain the KV cache for the full context. At each generation step $t$, the Indexer performs an attention computation over the preceding $t-1$ tokens, selecting $M$ important tokens to fill into the self-attention, which together with the linear attention state at step $t$ produces the $t$-th token. The resulting state is used to update both the linear attention state and the KV cache.

The Indexer's role is to select important tokens from the full context for subsequent token generation, which is similar to DeepSeek V3.2~\citep{deepseek2025v32} and GLM-5~\citep{zai2026glm5}. The key difference is that, due to the top-$M$ (argmax) strategy, training gradients cannot flow back to the Indexer through the discrete selection. Their approach maintains a distillation loss to update the Indexer; V3.2 in particular continues from a prior checkpoint with a backbone warm-up rather than training from a randomly initialized LLM, which differs from our post-hoc conversion of a frozen reasoning checkpoint. Building on this, we introduce a candidate-set dropout as regularization: during early training we take the top-$\lceil 1.25M\rceil$ scored positions and uniformly subsample exactly $M$ of them to fill the self-attention stream, forcing the model to make robust selections among near-top candidates. As training progresses, we gradually shrink the candidate pool toward $M$. This dropout regularizes the selected set and the downstream sparse branch; it does \emph{not} restore gradients through the hard top-$M$ gate---the Indexer parameters are updated by the KL objective below.

However, this still limits the model's performance to some extent. Therefore, we introduce a \textbf{multi-head mechanism} for the Indexer, with the loss applied directly on each head. We set the Indexer width to $d_I{=}256$ and assign one Indexer head to every four teacher attention heads, i.e.\ $K{=}P/4$ where $P$ is the number of teacher heads (yielding $K{=}3$ on the 1.5B model with $P{=}12$ and $K{=}7$ on the 7B model with $P{=}28$). Each Indexer head is trained to fit the attention distribution obtained by summing and re-normalizing its four assigned teacher heads.

Concretely, we sum the scores of $K$ Indexer heads over all tokens and select the $M$ positions with the largest values as \textbf{important tokens}. Based on the Indexer's selection, we fill the cached hidden states at the corresponding positions into $\mathrm{SA\_M}$ for subsequent computation. This approach mitigates the loss of precise contextual information caused by linear attention state compression.

The linear attention continues to be updated via cross-entropy loss. The Indexer loss is defined as:
\begin{equation}
\mathcal{L}_{\mathrm{Indexer}} = \frac{1}{KL} \sum_{j=1}^{K} \sum_{t=1}^{L} \mathrm{KL}\!\left(T^{(j)}_{t,:} \,\Big\|\, \mathrm{softmax}(S^{(j)}_{t,:})\right),
\end{equation}
where $S^{(j)}_{t,:} \in \mathbb{R}^{L}$ is the ReLU dot-product score vector of the $j$-th Indexer head over all keys, computed as $\mathrm{ReLU}\bigl(\langle q^{(j)}_t,\, k^{(j)}_{\cdot}\rangle\bigr)$, with $q^{(j)}_t = \mathrm{RoPE}(W^Q_j h_t)$ and $k^{(j)}_{\cdot} = \mathrm{RoPE}(W^K_j h_{\cdot})$. The target distribution $T^{(j)}_{t,:}$ is the normalized teacher target distribution for the $j$-th Indexer head, obtained by summing the attention distributions of its four assigned teacher heads and re-normalizing (with stop-gradient applied). Both $\mathrm{softmax}(\cdot)$ and the normalization in $T^{(j)}_{t,:}$ are restricted to the causal triangle for row-wise normalization.

\subsection{Inference State Transfer}

During inference, leveraging the test-time training (TTT) properties inherent in linear attention, we apply state-transfer operations to stabilize the model's reasoning trajectory. For segment-level correction we follow~\citet{zhang2026statetransitionframeworkefficientllm} (Liang~Zhang et~al.), partitioning generation into fixed-length segments and applying their dual-update rule to synchronize linear-attention (LA) state transitions with segment-level corrections. Our contributions relative to that line of work are the dynamic Indexer, per-head Indexer supervision, and their combination with this adopted correction rule.

\paragraph{LA State Update.}
In the $l$-th layer LA submodule, after completing the first $t-1$ reasoning segments, the model state is updated to $\mathbf{S}_{t-1}^{(l)}$. This serves as the initial state $\mathbf{S}_{t,0}^{(l)} = \mathbf{S}_{t-1}^{(l)}$ for the current reasoning segment and is updated token-by-token:
\begin{equation}
    \mathbf{S}_{t,i-1}^{(l)} = \mathbf{S}_{t,0}^{(l)} + \sum_{j=1}^{i-1} {\mathbf{k}_{j}^{(l)}}^{\top} \mathbf{v}_{j}^{(l)}.
\end{equation}
We then use the query vector $\mathbf{q}_{i-1}^{(l)}$ of input token $\mathbf{x}_{i-1}$ to extract relevant historical information from this state:
\begin{equation}
    \mathbf{o}_{i-1}^{(l)} = \mathbf{q}_{i-1}^{(l)} \mathbf{S}_{t,i-1}^{(l)}.
\end{equation}
We obtain the submodule output via a gating mechanism:
\begin{equation}
\check{\mathbf{h}}_{t,i-1}^{(l)} = \sigma\!\left(\mathbf{w}_g^{\top} \mathbf{h}_{t,i-1}^{(l-1)}\right) \odot \mathbf{o}_{i-1}^{(l)},
\end{equation}
where $\sigma$ is the sigmoid and $\mathbf{w}_g \in \mathbb{R}^{d}$ is a lightweight per-layer gate (a scalar after the sigmoid, broadcast over channels). The LA submodule is implemented via a LoRA~\citep{hu2022lora} strategy with rank $r{=}256$.

\paragraph{Inference-Time Segment Correction.}
Noisy reasoning segments can push the linear-attention state off a stable trajectory. We adopt the segment correction of~\citet{zhang2026statetransitionframeworkefficientllm} without claiming it as a contribution: partition generation into fixed-size segments, take each segment's state change as its reasoning direction, and pull it toward the momentum-averaged direction of preceding segments before applying the update. A brief restatement is given in Appendix~\ref{app:segment}; the contribution of applying this rule in our setting is isolated in Table~\ref{tab:ablation} (\textit{No segment correction}, Avg 50.4 vs.\ 57.1).
\section{Experiments}\label{sec:exps}

\subsection{Implementation Details}

Our experiments utilize the Qwen-2.5 model series~\citep{qwen2025qwen25technicalreport}. To construct the training set, we randomly sample 100K samples from the OpenR1-Math-220K dataset. The model is initialized using the weights of its corresponding DeepSeek-R1 distilled version~\citep{deepseek2025r1}.

During Stage~1 of training, we employ a learning rate of $2\times10^{-5}$ and a batch size of 128 for 2 epochs with $M=256$. To initialize the linear attention layers, we evaluated both random initialization and Singular Value Decomposition (SVD) of the base model's self-attention weights. Since SVD initialization yielded no measurable performance gains, all reported experiments utilize random initialization accompanied by a 50-step linear warmup. This stage takes approximately 240 GPU hours on an NVIDIA H100. The Indexer uses per-head width $d_I{=}256$ with one Indexer head per four teacher heads ($K{=}P/4$; $K{=}3$ on 1.5B and $K{=}7$ on 7B). Linear-attention adapters are implemented with LoRA~\citep{hu2022lora} of rank $r{=}256$. Full attention is retained at blocks $0,7,14,21,27$ (5 of 28); the remaining 23 blocks are LISA layers.

For Stage~2, the model is trained for 1 epoch using a learning rate of $2\times10^{-5}$ and a batch size of 64 with $M=256$. We initialize the Indexer randomly with a 50-step warmup. To mitigate the impact of initial noisy Indexer representations, the linear attention module is frozen for the first 200 steps. Subsequently, the Indexer and linear attention are updated simultaneously. With a halved batch size relative to Stage~1, the number of optimizer steps is comparable, but wall-clock cost rises substantially because each step requires teacher attention computation, taking approximately 240 GPU hours on an NVIDIA H100.

\paragraph{Parameter and Memory Overhead.}
LISA retains the full KV cache, so the stored KV footprint is not reduced by the proposed method. Beyond the lightweight auxiliary parameters (Linear Attention with LoRA rank $r{=}256$, the Indexer with $d_I{=}256$, and a per-layer scalar gate)---roughly 111M extra parameters (+7.4\%) for the 1.5B model and 420M (+6.0\%) for the 7B model---LISA also maintains recurrent LA states and an Indexer-key cache that grows linearly with context length. The bulk of the parameter overhead comes from the LA adapters and Indexer; the gate contributes only a small residual. The auxiliary \emph{parameter} count itself is fixed and independent of context length.

\subsection{Baselines}

We compare our framework with two types of baselines: (1) \textbf{Efficient reasoning methods}: \textbf{LightThinker}~\citep{zhang2025lightthinker} uses learnable special tokens to compress the reasoning information of completed reasoning steps; \textbf{INFTYTHINK}~\citep{yan2025inftythink} compresses prior reasoning steps into a concise summary; \textbf{\citet{zhang2026statetransitionframeworkefficientllm}} proposes a state-transition framework that uses linear attention to maintain a global state matrix for recording historical information, enabling the model to retrieve relevant information without explicitly attending to prior reasoning steps. (2) \textbf{KV-cache reduction methods}: \textbf{H2O}~\citep{zhang2023h2o} retains only the KV vectors of heavy-hitter tokens and recent tokens; \textbf{SepLLM}~\citep{chen2025sepllm} only stores the KV vectors of initial, neighboring, and separator tokens. All baselines are trained using the LoRA~\citep{hu2022lora} strategy with a comparable number of trainable parameters to our model.

\subsection{Datasets \& Metrics}\label{subsec:dataset_metric}

We evaluate our framework on five mathematical benchmarks (GSM8K~\citep{cobbe2021verifiers}, MATH-500~\citep{hendrycks2021math}, AMC23~\citep{amc2023}, AIME24~\citep{aime2024} \& 25~\citep{aime2025}), and assess generalization on four non-mathematical benchmarks (GPQA-D~\citep{rein2024gpqa}, HumanEval~\citep{chen2021humaneval}, LiveCodeBench~\citep{jain2024livecodebenchholisticcontaminationfree}, and BigCodeBench~\citep{zhuo2025bigcodebenchbenchmarkingcodegeneration}). We report three metrics: (1) \textit{Accuracy} (\textbf{Acc}) denotes the percentage of correct answers; (2) \textit{Token Number} (\textbf{Tok}) refers to the average length of the CoT sequence; (3) \textit{Reasoning Latency} (\textbf{ReL}) is defined as the average wall-clock inference time per sample (seconds). We measure reasoning latency on a single NVIDIA A100 GPU, using eager mode with BF16 precision.

Note that latency measured in eager mode does not reflect the actual speedup attainable under production-grade inference frameworks such as vLLM or SGLang, whose advantages stem largely from kernel-level optimizations (e.g., PagedAttention, FlashInfer). Integrating LISA's linear attention, indexed sparse attention, and gating into such frameworks requires custom kernel development and is beyond the scope of this work. A secondary decoding-FLOPs estimate that includes Indexer scoring is given in Appendix~\ref{app:flops}; our primary efficiency claim is the measured ReL.

In addition, since AMC23, AIME24, and AIME25 contain only about 30 problems each, we evaluate these three benchmarks on the Qwen2.5-1.5B model over 16 independent runs and report the mean $\pm$ standard deviation in Table~\ref{tab:variance}. Table~\ref{tab:main_math} itself reports single representative runs throughout.

\begin{table*}[t]
    \centering
    \resizebox{\textwidth}{!}{%
    \begin{tabular}{l ccc ccc ccc ccc ccc cc}
    \toprule
    \multirow{2}{*}{Method} & \multicolumn{3}{c}{GSM8K} & \multicolumn{3}{c}{MATH-500} & \multicolumn{3}{c}{AMC23} & \multicolumn{3}{c}{AIME24} & \multicolumn{3}{c}{AIME25} & \multicolumn{2}{c}{AVG.} \\
    \cmidrule(lr){2-4} \cmidrule(lr){5-7} \cmidrule(lr){8-10} \cmidrule(lr){11-13} \cmidrule(lr){14-16} \cmidrule(lr){17-18}
     & Acc\textuparrow & Tok\textdownarrow & ReL\textdownarrow & Acc\textuparrow & Tok\textdownarrow & ReL\textdownarrow & Acc\textuparrow & Tok\textdownarrow & ReL\textdownarrow & Acc\textuparrow & Tok\textdownarrow & ReL\textdownarrow & Acc\textuparrow & Tok\textdownarrow & ReL\textdownarrow & Acc\textuparrow & ReL\textdownarrow \\
    \midrule
    \multicolumn{18}{c}{\textit{Qwen2.5-1.5B Series}} \\
    \midrule
    Full-attention baseline & 80.1 & 2086 & 76.2 & 78.8 & 3958 & 139.4 & 62.5 & 6392 & 242.9 & 20.0 & 13765 & 536.8 & 16.7 & 12684 & 504.7 & 51.6 & 300.0 \\
    \quad + H2O & 76.1 & 2116 & 71.7 & 75.4 & 3835 & 128.5 & 55.0 & 6230 & 209.7 & 13.3 & 13921 & 462.2 & 10.0 & 12802 & 428.6 & 46.0 & 260.1 \\
    \quad + SepLLM & 77.3 & 2230 & 75.8 & 76.0 & 3890 & 130.3 & 57.5 & 6573 & 220.2 & 13.3 & 13690 & 452.6 & 13.3 & 12690 & 423.9 & 47.5 & 260.6 \\
    LightThinker & 78.8 & 2109 & 69.6 & 77.6 & 4060 & 134.8 & 60.5 & 6312 & 214.6 & 16.7 & 13827 & 461.6 & 13.3 & 12934 & 434.6 & 49.4 & 263.0 \\
    INFTYTHINK & 79.5 & 2503 & 89.3 & 78.0 & 4750 & 170.2 & 62.5 & 7605 & 263.5 & 16.7 & 16318 & 566.2 & 16.7 & 15020 & 501.0 & 50.7 & 318.0 \\
    \citet{zhang2026statetransitionframeworkefficientllm} & 80.1 & 2116 & 69.7 & 81.2 & 3812 & 125.8 & 67.5 & 6460 & 213.2 & 26.7 & 13610 & 440.1 & 23.3 & 12910 & 416.0 & 55.8 & 253.0 \\
    \textbf{LISA} & \textbf{80.2} & \textbf{1983} & \textbf{61.5} & \textbf{82.0} & \textbf{3636} & \textbf{102.5} & \textbf{70.0} & \textbf{6157} & \textbf{187.7} & \textbf{26.7} & \textbf{12966} & \textbf{392.3} & \textbf{26.7} & \textbf{12532} & \textbf{398.1} & \textbf{57.1} & \textbf{228.4} \\
    \midrule
    \multicolumn{18}{c}{\textit{Qwen2.5-7B Series}} \\
    \midrule
    Full-attention baseline & 89.4 & 2649 & 96.4 & 87.4 & 4253 & 161.6 & 82.5 & 6704 & 265.2 & 40.0 & 12901 & 522.1 & 30.0 & 13204 & 548.1 & 65.9 & 318.7 \\
    \quad + H2O & 83.5 & 2730 & 92.0 & 82.6 & 4389 & 147.1 & 77.5 & 6843 & 221.5 & 30.0 & 13274 & 448.9 & 23.3 & 13204 & 443.7 & 59.4 & 270.6 \\
    \quad + SepLLM & 84.9 & 2582 & 92.3 & 84.2 & 4244 & 144.3 & 80.0 & 6511 & 231.2 & 33.3 & 13100 & 484.7 & 23.3 & 12972 & 435.9 & 61.1 & 277.7 \\
    LightThinker & 85.6 & 2779 & 90.0 & 84.8 & 4321 & 145.2 & 80.0 & 6790 & 242.5 & 33.3 & 13307 & 449.0 & 26.7 & 13250 & 443.9 & 62.1 & 274.1 \\
    INFTYTHINK & 87.9 & 3237 & 117.0 & 86.2 & 5050 & 184.8 & 82.5 & 8061 & 261.9 & 36.7 & 15094 & 508.7 & 30.0 & 15448 & 530.5 & 64.7 & 320.6 \\
    \citet{zhang2026statetransitionframeworkefficientllm} & 90.9 & 2603 & 87.2 & 90.0 & 4196 & 140.6 & 85.0 & 6665 & 245.1 & 43.3 & 13017 & 434.8 & 33.3 & 13191 & 440.6 & 68.5 & 269.7 \\
    \textbf{LISA} & \textbf{90.9} & \textbf{2462} & \textbf{82.1} & \textbf{91.2} & \textbf{4253} & \textbf{117.9} & \textbf{87.5} & \textbf{6841} & \textbf{196.3} & \textbf{43.3} & \textbf{12587} & \textbf{385.1} & \textbf{40.0} & \textbf{13642} & \textbf{398.1} & \textbf{70.6} & \textbf{235.9} \\
    \bottomrule
    \end{tabular}
    }%
    \caption{Results of our model and baselines on mathematical reasoning benchmarks. The \textit{Full-attention baseline} row is the unmodified original model. All cells report a single representative run; run-to-run variance on the small-sample benchmarks is reported separately in Table~\ref{tab:variance}.}
    \label{tab:main_math}
\end{table*}

\begin{table}[t]
    \centering
    \resizebox{0.6\textwidth}{!}{%
    \begin{tabular}{l cc cc cc}
    \toprule
     & \multicolumn{2}{c}{AMC23} & \multicolumn{2}{c}{AIME24} & \multicolumn{2}{c}{AIME25} \\
    \cmidrule(lr){2-3} \cmidrule(lr){4-5} \cmidrule(lr){6-7}
     & Acc\textuparrow & $\pm$std & Acc\textuparrow & $\pm$std & Acc\textuparrow & $\pm$std \\
    \midrule
    Full-attention baseline & 62.5 & 2.24 & 19.6 & 1.65 & 16.1 & 1.83 \\
    \textbf{LISA} & \textbf{70.0} & 2.74 & \textbf{25.8} & 1.96 & \textbf{25.9} & 2.29 \\
    \bottomrule
    \end{tabular}
    }%
    \caption{Mean accuracy ($\%$) over 16 independent evaluation runs on the small-sample benchmarks, Qwen2.5-1.5B. These benchmarks contain only about 30 problems each, so we report the standard deviation alongside the mean; the corresponding single-run values in Table~\ref{tab:main_math} fall within one standard deviation of these means.}
    \label{tab:variance}
\end{table}

\subsection{Main Results}

We conduct a comprehensive evaluation of LISA across five mathematical reasoning benchmarks. The results, summarized in Table~\ref{tab:main_math}, compare LISA against state-of-the-art efficiency-driven baselines using both Qwen2.5-1.5B and Qwen2.5-7B backbones.

\paragraph{LISA improves average accuracy and efficiency.}
LISA attains the highest average accuracy on the mathematical suite and matches or exceeds the compared methods on each benchmark, while reporting the lowest mean per-sample latency (\textbf{ReL}$\downarrow$) among all tested methods. On the Qwen2.5-7B series, LISA reaches an average accuracy of 70.6\%, outperforming the strongest baseline~\citep{zhang2026statetransitionframeworkefficientllm} by 2.1~percentage points, while reducing mean ReL from 269.7 to 235.9. Because the full-attention baseline is the unmodified checkpoint without our conversion training, accuracy gains should be read as the combined effect of the hybrid architecture and the additional Stage~1/2 training rather than as a pure architectural delta. The 23.9\% ReL reduction quoted in the abstract is this table's 1.5B average (300.0$\to$228.4) and is not a fixed-16K per-token measurement; generation lengths also differ slightly across methods (\# Tok), so latency comparisons inherit that confound.

\paragraph{Impact of sequence length and task complexity.}
A key observation from Table~\ref{tab:main_math} is the correlation between task complexity (and output length) and the performance margin of our method. Specifically, on benchmarks with short outputs like GSM8K, the performance gap between LISA and the full-attention baseline is minimal (e.g., 80.2\% vs 80.1\% for the 1.5B model). The primary reason is that our method still introduces a fixed budget for self-attention, yielding marginal gains on short sequences. By contrast, on tasks requiring long-horizon reasoning, the advantages of LISA become more pronounced. For instance, on AIME25 with the 7B model, LISA achieves 40.0\% accuracy, significantly higher than the 33.3\% reached by~\citet{zhang2026statetransitionframeworkefficientllm}. This suggests that our sparse attention mechanism is particularly adept at maintaining coherence and reasoning accuracy in long-sequence generation where standard KV-cache or attention-reduction methods often struggle.

\subsection{Generalization to Non-Mathematical Tasks}

To verify that LISA generalizes beyond mathematical reasoning, we also evaluate the Qwen2.5-1.5B model on science (GPQA-D~\citep{rein2024gpqa}) and code domains (HumanEval~\citep{chen2021humaneval}, LiveCodeBench~\citep{jain2024livecodebenchholisticcontaminationfree}, and BigCodeBench~\citep{zhuo2025bigcodebenchbenchmarkingcodegeneration}); the results are reported in Table~\ref{tab:generalization}. LISA preserves baseline performance on these non-mathematical tasks, indicating that the architectural conversion does not introduce domain-specific degradation.

\begin{table}[t]
    \centering
    \resizebox{0.8\textwidth}{!}{%
    \begin{tabular}{lcccc}
    \toprule
    Method & GPQA-D\textuparrow & HumanEval\textuparrow & LiveCodeBench\textuparrow & BigCodeBench\textuparrow \\
    \midrule
    Full-attention baseline & 23.23 & 76.22 & 35.0 & 22.7 \\
    \textbf{LISA} & \textbf{26.26} & \textbf{79.27} & \textbf{37.4} & \textbf{24.7} \\
    \bottomrule
    \end{tabular}
    }
    \caption{Generalization results of Qwen2.5-1.5B on non-mathematical tasks (accuracy, \%).}
    \label{tab:generalization}
\end{table}

\subsection{Component-Level Ablation}

We conduct a full component-level ablation on DeepSeek-R1-Distill-Qwen-1.5B, with all variants sharing the same training configuration; the results are reported in Table~\ref{tab:ablation}. Removing either branch (Linear only / Sparse only) causes a substantial drop in performance, indicating that the two branches each provide information the other cannot substitute. Moreover, replacing Indexer-guided selection with a sliding window (Stage 1 only) also degrades performance; this variant additionally omits Stage~2 training, so the gap reflects both the selection mechanism and the missing Stage~2 updates. Among the remaining components the ordering is consistent: removing segment correction costs $-6.7$ Avg, more than gating ($-3.7$) and the multi-head Indexer ($-1.5$) combined. Segment correction is therefore the largest of the three despite being the only one that adds no learnable parameter---it acts purely at inference time. These results support the effectiveness of segment correction; linking it causally to ``overthinking'' would require more direct measurements (e.g., error-continuation or state-drift statistics) and is left for future analysis.

\begin{table}[t]
    \centering
    \resizebox{\textwidth}{!}{%
    \begin{tabular}{ll ccccc c}
    \toprule
    \multirow{2}{*}{Variant} & \multirow{2}{*}{Description} & GSM8K & MATH-500 & AMC23 & AIME24 & AIME25 & \multirow{2}{*}{Avg} \\
    \cmidrule(lr){3-7}
     & & Acc\textuparrow & Acc\textuparrow & Acc\textuparrow & Acc\textuparrow & Acc\textuparrow & \\
    \midrule
    \textbf{Full LISA (ours)} & Complete method (Stage 1 + Stage 2) & \textbf{80.2} & \textbf{82.0} & \textbf{70.0} & \textbf{26.7} & \textbf{26.7} & \textbf{57.1} \\
    Stage 1 only & Sliding-window sparse without indexer-guided training & 79.2 & 80.6 & 65.0 & 23.3 & 23.3 & 54.3 \\
    Linear only & Remove sparse branch & 68.3 & 67.8 & 55.0 & 13.4 & 10.0 & 42.9 \\
    Sparse only & Remove linear branch & 66.6 & 60.8 & 42.5 & 6.7 & 6.7 & 36.7 \\
    No gating & Remove learned gating & 78.3 & 80.2 & 65.0 & 23.3 & 20.0 & 53.4 \\
    Single-head Indexer & Replace multi-head with single-head Indexer & 79.5 & 81.2 & 67.5 & 26.7 & 23.3 & 55.6 \\
    No segment correction & Remove $k$-step correction at inference & 74.9 & 77.8 & 62.5 & 20.0 & 16.7 & 50.4 \\
    \bottomrule
    \end{tabular}
    }%
    \caption{Component-level ablation on Qwen2.5-1.5B (accuracy, \%). All variants use the same training configuration.}
    \label{tab:ablation}
\end{table}

\paragraph{Framework-agnostic efficiency measure.}
Since latency measured in eager mode does not track production inference frameworks, we additionally report a secondary per-token decoding-FLOPs estimate in Table~\ref{tab:flops} (Appendix~\ref{app:flops}), including Indexer full-context scoring. Gains grow with context length and remain moderate; our primary efficiency claim is the measured ReL reduction above. After selection, sparse self-attention in a LISA layer costs $O(M)$ per decoded token, while Indexer scoring remains linear in the current context length and retained full-attention layers remain $O(t)$, so the overall sequence-level complexity stays hybrid / quadratic.

\subsection{Analysis}

\begin{figure}[tp]
    \centering
    \setlength{\abovecaptionskip}{-0.2cm}
    \setlength{\belowcaptionskip}{-0.3cm}
    \includegraphics[width=0.72\textwidth]{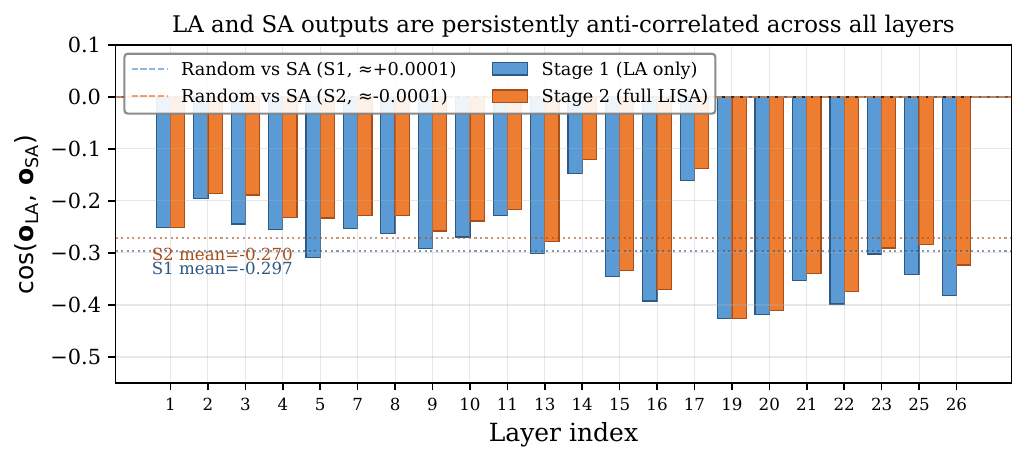}
    \caption{Cosine similarity between the linear and sparse attention outputs across all layers.}
    \label{fig:la_sa_cosine}
\end{figure}

\paragraph{Analyzing the Synergy between Linear and Sparse Attention.}
To verify that the linear branch actually supplies the global context it is intended to provide, we compare the hidden states of the two branches by computing the cosine similarity between their outputs (Figure~\ref{fig:la_sa_cosine}). The similarity is consistently negative across all layers, indicating that the two representations are non-trivially anti-aligned in direction rather than near-duplicates.

A negative similarity is by itself only a correlational observation, however. The causal evidence is the component-level ablation in Table~\ref{tab:ablation}: removing either branch degrades performance sharply, and the Stage~1-only sliding-window variant also underperforms full LISA. Consistently, the similarity shifts from $-0.297$ to $-0.270$ in Stage~2,\footnote{These magnitudes are small in absolute value; for scale, the cosine similarity between random vectors and sparse attention outputs is approximately $10^{-4}$, three orders of magnitude lower.} which indicates that as the Indexer widens the sparse branch's receptive field, the two signals converge partially. A companion analysis of what the Indexer itself selects across layers is provided in Appendix~\ref{app:indexer}.

\paragraph{Window Size $M$ and Attention Coverage.}
We investigate the impact of $M$ on both sliding-window size and Indexer capacity. To quantify coverage of the teacher's attention mass, we define:
\begin{equation}
    \textsc{mass}(q, M) = \frac{\sum_{k \in \text{top-}M(q)} \text{Attn}(q, k)}{\sum_{\text{all } k \leq q} \text{Attn}(q, k)},
\end{equation}
where $\text{Attn}(q, k)$ denotes the \emph{teacher} attention weights. When evaluating the Indexer, $\text{top-}M(q)$ are the positions selected by the Indexer; the oracle upper bound instead takes the teacher's own top-$M$ positions. At $M=256$, Indexer \textsc{mass} reaches 80\%, trailing the oracle upper bound by 6~percentage points (86\%). The metric is 70\% at $M=128$ and 88\% at $M=512$. We adopt $M=256$ as the default throughout our experiments as a practical coverage--cost trade-off; establishing optimality would additionally require paired task-accuracy and latency curves, which we do not claim here.
\section{Conclusion}

We proposed LISA (Linear-Indexed Sparse Attention), a plug-and-play attention mechanism that accelerates long-context chain-of-thought reasoning without pre-training from scratch. It runs two parallel streams --- a global linear attention branch for long-range context and a fixed-size sparse self-attention branch for high-fidelity local and salient reasoning --- trained in two stages: cold-starting linear attention for long-range dependencies, then introducing an indexer for dynamic token selection. This replaces dense attention in each LISA layer with attention over a fixed budget of $M$ selected tokens ($M \ll n$); because full-attention layers are retained periodically and Indexer scoring still scans the full context, the overall sequence-level complexity class remains quadratic. On mathematical reasoning and on non-mathematical tasks including science question answering and code generation, LISA reduces measured inference latency (23.9\% mean ReL on 1.5B) while matching or surpassing competitive baselines. Future work includes adaptive token budgets, integration into production inference frameworks for end-to-end speedups, and long-context retrieval and agent-based settings.

\subsection*{AI use statement}

Generative AI was not used to implement LISA, design or run experiments, interpret results, formulate mathematical claims, propose hypotheses, or generate, clean, or
reformat datasets. The scientific contributions, experimental conclusions, and final claims were determined and independently verified by the authors. 

Generative AI tools were exclusively used to polish the grammar and phrasing of this paper. The authors checked every AI-assisted passage against the original manuscript, experimental records, source code, and cited references. They also manually verified every equation, numerical claim, and citation in the paper. AI-suggested wording or structure was accepted only when it preserved the intended scientific meaning and introduced no unsupported claim. The authors take responsibility for the final content, including text, claims, and artifacts produced with AI assistance.

\subsection*{Ethics statement}






This work studies inference efficiency and involves no human participants, personal data, or new data collection. Training and evaluation use the public datasets and benchmarks described in Section~\ref{subsec:dataset_metric}. These resources are used under their published terms, and the authors will not redistribute dataset content beyond what those terms permit. Any release will contain only code, model components, and configurations.

LISA may reduce the compute required to serve long-reasoning models, which could improve
access and lower per-query resource use. The same efficiency gain could also reduce the cost of deploying capable models for harmful purposes. LISA adds no safety mechanism and does not remove biases or other risks inherited from the base models; deployments should retain the evaluation, access controls, and monitoring appropriate for those models.

To support research integrity, we report the retained quadratic components, additional training cost, evaluation scope, unmatched training budget, and run-to-run variance instead of presenting the method as an asymptotic or universally applicable improvement. Training the added components also incurs substantial GPU cost, which should be weighed against any inference savings. The method changes the attention mechanism rather than the model's safety properties. No additional privacy, human-subjects, or discrimination concerns arise beyond those associated with the public data and pretrained models used in this study

\subsection*{Reproducibility statement}


Section~\ref{sec:method} specifies the LISA architecture, objectives, token-selection rule, gating, and inference-time segment correction in closed form. Section~\ref{sec:exps} gives the base checkpoints used, the 100K-sample training set, the complete two-stage optimization schedule, and the evaluation and latency protocols. It reports learning rates, batch sizes, epochs, warmups, the freezing schedule, LoRA rank, Indexer dimensions, retained-attention layers, token budget, hardware, metrics, and precision. Tables 1-4 report the main results, variance, and ablations, while the appendix provides the FLOP analysis and complete token-selection traces.




\bibliography{iclr2027_conference}
\bibliographystyle{iclr2027_conference}
\appendix

\section{Decoding FLOPs Analysis}
\label{app:flops}

Table~\ref{tab:flops} reports a secondary, framework-agnostic estimate of per-token \textbf{decoding FLOPs} (attention block + SwiGLU MLP; excluding the embedding and LM-head). Our primary efficiency claim remains the measured eager-mode latency (ReL) in the main experiments. After selection, sparse self-attention in a LISA layer costs $O(M)$ per decoded token; Indexer scoring remains linear in the current context length and retained full-attention layers remain $O(t)$, so the overall sequence-level complexity is hybrid / quadratic and any FLOPs gain is a constant factor.

\paragraph{Attention block.}
The attention FLOPs of a single layer for one decoded token are
\begin{equation}
\mathrm{FLOPs}_{\mathrm{attn}}
=
\underbrace{2d(d+2d_{kv})}_{\mathrm{QKV}}
+
\underbrace{4\,n_h\,d_h\,L}_{\mathrm{QK}^{\top}+\mathrm{AV}}
+
\underbrace{2d^{2}}_{\mathrm{O}},
\end{equation}
where $d_{kv}=n_{kv}d_h$ and $L$ is the number of keys attended in that layer. The full-attention baseline uses $L{=}t$ in all 28 layers. For LISA we use $L{=}t$ in the 5 retained full-attention blocks ($0,7,14,21,27$) and $L{=}M{=}256$ in the sparse branch of the remaining 23 LISA layers, and we \emph{add} linear-attention projections/state updates plus Indexer scoring with cached keys ($d_I{=}256$, $K{=}P/4$). With a key cache, Indexer cost per LISA layer is $4dKd_I + 2Kd_It$ (query/key projections for the new token and ReLU scores against $t$ cached keys). Substituting the Qwen2.5-1.5B configuration ($d{=}1536$, $n_h{=}12$, $n_{kv}{=}2$, $d_h{=}128$, $K{=}3$) at $t{=}16$K yields $3.13$\,GFLOPs over 28 full-attention layers for the baseline attention component before MLP.

\paragraph{MLP.}
Qwen2.5 uses SwiGLU, so each layer costs three projections per token,
\begin{equation}
\mathrm{FLOPs}_{\mathrm{MLP}} = 6\,d\,d_{\mathrm{ff}},
\end{equation}
times 28 layers. LISA leaves the MLP unchanged: $2.31$\,GFLOPs for the 1.5B model ($d_{\mathrm{ff}}{=}8960$) and $11.41$\,GFLOPs for the 7B model ($d{=}3584$, $d_{\mathrm{ff}}{=}18944$).

\paragraph{Total and Table~\ref{tab:flops}.}
Per-token decoding FLOPs ${=} \mathrm{FLOPs}_{\mathrm{SA}}+\mathrm{FLOPs}_{\mathrm{Indexer}}+\mathrm{FLOPs}_{\mathrm{LA}}+\mathrm{FLOPs}_{\mathrm{MLP}}$. For 1.5B at 16K this gives full attention $5.44$\,G versus LISA $4.01$\,G (about $26\%$ reduction); for 7B the corresponding figure is about $15\%$. Savings are small or negligible at short context and grow with $t$, as expected once Indexer scoring is counted. The measured 23.9\% ReL cut on 1.5B mathematical benchmarks is therefore the headline efficiency result; the table is only a rough upper-bound check under the accounting above.

\begin{table}[t]
    \centering
    \begin{tabular}{l ccc ccc}
    \toprule
    \multirow{2}{*}{Context} & \multicolumn{3}{c}{Qwen2.5-1.5B} & \multicolumn{3}{c}{Qwen2.5-7B} \\
    \cmidrule(lr){2-4} \cmidrule(lr){5-7}
     & Full-Attn & LISA & Reduction & Full-Attn & LISA & Reduction \\
    \midrule
    4K  & 3.33 G & 3.20 G & 3.8\% & 14.69 G & 14.74 G & $-$0.3\% \\
    8K  & 4.03 G & 3.47 G & 13.9\% & 16.34 G & 15.37 G & 5.9\% \\
    12K & 4.73 G & 3.74 G & 21.0\% & 17.98 G & 16.00 G & 11.0\% \\
    16K & 5.44 G & 4.01 G & 26.3\% & 19.63 G & 16.63 G & 15.3\% \\
    \bottomrule
    \end{tabular}
    \caption{Secondary decoding-FLOPs estimate per generated token (GFLOPs; SA + Indexer with cached keys + LA + SwiGLU MLP). LISA uses 5 full-attention + 23 LISA blocks. Primary efficiency claim: measured ReL.}
    \label{tab:flops}
\end{table}

\section{Indexer Output Distribution}
\label{app:indexer}

\begin{figure}[ht]
    \centering
    \includegraphics[width=0.7\textwidth]{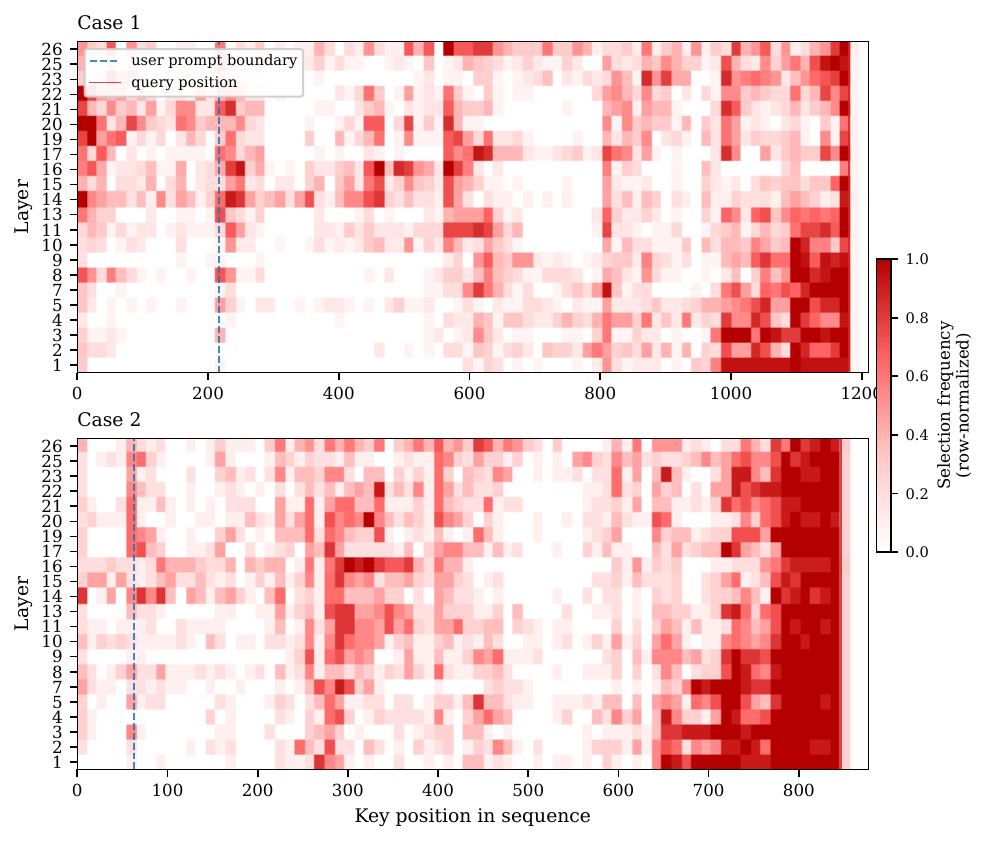}
    \caption{Visualization of the top-256 tokens selected by the Indexer across layers.}
    \label{fig:indexer_distribution}
\end{figure}

\paragraph{Indexer Selection Patterns.}
We conduct an analysis of the Indexer's outputs to check alignment with known sparse-attention behaviors. By visualizing the top-256 tokens selected by the Indexer for two randomly sampled examples (Figure~\ref{fig:indexer_distribution}), we observe that the mechanism identifies locally recent tokens as well as some global landmarks.

Specifically, the distribution of selected tokens is heavily concentrated in the local context of the current output token, consistent with recency bias in transformers~\citep{peysakhovich2023attention, wang2024eliminating, xiao2023streamingllm}. Beyond the local window, the Indexer selectively attends to global features such as punctuation tokens, in line with prior sparse-attention observations~\citep{lou2024sparser, bondarenko2023quantizable}, and attention-sink behavior~\citep{xiao2023streamingllm} is visible. These qualitative case studies are illustrative rather than a proof that the Indexer recovers the full teacher attention distribution; quantitative agreement (e.g., top-$M$ overlap or KL to the teacher) over a larger sample is left for future work.

The visualisations below repeat this analysis per layer: they show the top-256 keys selected by the LISA Indexer for the same two reasoning examples at different network depths, where the background colour intensity (white $\to$ dark red) reflects the attention score assigned to each candidate key token.

\subsection{Layer~1 Indexer Output Distribution}

\begin{tcolorbox}[
    enhanced, breakable,
    colback=white, colframe=black, boxrule=0.5pt,
    arc=2pt, left=4pt, right=4pt, top=4pt, bottom=4pt,
    fonttitle=\bfseries,
    width=\linewidth,
]
{\raggedright \ttfamily \footnotesize
    \textbf{Question:} \colorbox[HTML]{FFFDFD}{\strut <|begin of sentence|>}\colorbox[HTML]{FFF8F8}{\strut <|User|>}The center of a circle touching the legs \$A C\$ and \$B C\$ of a right triangle \$A B C\$ lies on the hypotenuse \$A B\$. Find the radius of the circle, if it is six times smaller than the sum of the legs, and the area of triangle \$A B C\$ is 27.<|Assistant|> <think> \\
    
    \vspace{4pt} \hrule \vspace{4pt}
    \textbf{Response:} <think> \\
    Okay, so\colorbox[HTML]{FFFAFA}{\strut  I} have this problem about a right triangle ABC where the circle touches the legs AC and BC, and its center is on the hypotenuse AB. I need to find the radius of the circle, which is six times smaller than the sum of the legs. Also, the area of the triangle is 27. Hmm, let me break this down step by step. \\

    First, let me visualize the triangle. Since it's a right triangle, let's assume angle C is the right angle. So, legs AC and BC form the right angle, and AB is the hypotenuse. The circle touches both legs AC and BC, so it must be tangent to both. The center of the circle lies on AB. That makes me think it's an incircle or maybe an excircle, but since it's touching both legs and the center is on the hypotenuse, it's probably not the standard incircle because the incenter of a right triangle is located inside the triangle, not on the hypotenuse. Wait, but the problem says the center lies on the hypotenuse. So this must be a different circle, not the incircle. \\

    Let me recall that if a circle is tangent to both legs of a right triangle and its center is on the hypotenuse, there might be a specific formula or relation here. Let me denote the legs as AC = b and BC = a, and hypotenuse AB = c. The area is (a*b)/2 = 27, so a*b = 54. Also, the radius r is six times smaller than the sum of the legs, so r = (a + b)/6. My goal is to find r. \\

    Wait, so I need to relate r to a and b. Let me think about the coordinates. Maybe coordinate geometry can help here. Let me place point C at the origin (0,0), point B at (a,0), point A at (0,b). Then hypotenuse AB goes from (0,b) to (a,0). The center of the circle lies on AB and is equidistant from both legs AC and BC. Since the circle is tangent to both legs, which are the y-axis and x-axis respectively, the distance from the center to each axis must be equal to the radius. Wait, if the center is (r, r) for the incircle, but in this case, the center is on AB. So the center (h,k) must satisfy h = k = r? No, because the center is on AB. Wait, maybe the coordinates of the center can be param\colorbox[HTML]{FFFEFE}{\strut etr}\colorbox[HTML]{FFF5F5}{\strut ized} along the\colorbox[HTML]{FFF4F4}{\strut  hyp}otenuse. \\

    Let me param\colorbox[HTML]{FFE8E8}{\strut etrize} AB. The hypotenuse AB can be\colorbox[HTML]{FFFBFB}{\strut  represented} by the line from (0, b) to (a, 0). The equation of AB\colorbox[HTML]{FFFAFA}{\strut  is}\colorbox[HTML]{FFF7F7}{\strut  y}\colorbox[HTML]{FFF0F0}{\strut  =}\colorbox[HTML]{FFFCFC}{\strut  (-}b\colorbox[HTML]{FFEEEE}{\strut /a}\colorbox[HTML]{FFECEC}{\strut )x}\colorbox[HTML]{FFE8E8}{\strut  +} b.\colorbox[HTML]{FFEAEA}{\strut  Any}\colorbox[HTML]{FFF5F5}{\strut  point}\colorbox[HTML]{FFFAFA}{\strut  on} AB\colorbox[HTML]{FFF8F8}{\strut  can} be\colorbox[HTML]{FFEAEA}{\strut  written} as\colorbox[HTML]{FFFEFE}{\strut  (}\colorbox[HTML]{FFEBEB}{\strut x}\colorbox[HTML]{FFF7F7}{\strut ,}\colorbox[HTML]{FFFBFB}{\strut  (-}b\colorbox[HTML]{FFE9E9}{\strut /a}\colorbox[HTML]{FFEDED}{\strut )x}\colorbox[HTML]{FFE5E5}{\strut  +}\colorbox[HTML]{FFF3F3}{\strut  b}). The center of the circle is at some point (x, (-b/a)x + b) on AB, and this point must be at a distance r from both legs AC and BC. The legs AC and BC are the y-axis and x-axis respectively,\colorbox[HTML]{FFFFFF}{\strut  so} the distance from the center to AC (y-axis) is x, and the distance to BC (x-axis) is y. Since the circle is tangent to both legs, these distances must be equal to the radius. So x = r and y = r.\colorbox[HTML]{FFF5F5}{\strut  Therefore}, the center is (r, r), but\colorbox[HTML]{FFFDFD}{\strut  also} lies on AB.\colorbox[HTML]{FFFAFA}{\strut  Wait}, but if the center is (r, r), then\colorbox[HTML]{FFF3F3}{\strut  substit}\colorbox[HTML]{FFFBFB}{\strut uting} into the equation of AB gives r = (-b/a)r + b.\colorbox[HTML]{FFFBFB}{\strut  Let}\colorbox[HTML]{FFFEFE}{\strut  me}\colorbox[HTML]{FFEFEF}{\strut  check}\colorbox[HTML]{FFF5F5}{\strut  that}\colorbox[HTML]{FFFCFC}{\strut .} \\

    The equation of AB is y = (-b/a\colorbox[HTML]{FFFEFE}{\strut )x}\colorbox[HTML]{FFFAFA}{\strut  +} b.\colorbox[HTML]{FFEAEA}{\strut  So}\colorbox[HTML]{FFE5E5}{\strut  if}\colorbox[HTML]{FFF7F7}{\strut  the}\colorbox[HTML]{FFE6E6}{\strut  center} is at (r\colorbox[HTML]{FFFCFC}{\strut ,} r\colorbox[HTML]{FFF0F0}{\strut ),}\colorbox[HTML]{FFFAFA}{\strut  then} substituting x = r\colorbox[HTML]{FFFBFB}{\strut  into} the equation gives\colorbox[HTML]{FFFDFD}{\strut  y} = (-b/a)r + b. But the y-coordinate of the center is also r, so we have: \\

    r = (-b/a)r + b \\

    Let me solve\colorbox[HTML]{FFF9F9}{\strut  for} r: \\

    r + (b/a)r = b \\

    r(1 + b/a) = b \\

    r = b / (1 + b/a) = (ab) / (a + b) \\

    But from the problem statement, the radius r is (a + b)/6. So according to this, (ab)/(a + b) = (a + b)/6 \\

    \colorbox[HTML]{FFF0F0}{\strut Multiply} both\colorbox[HTML]{FFFBFB}{\strut  sides} by (a + b): \\

    ab = (a + b)\textasciicircum{}2 / 6 \\

    Multiply both sides by 6: \\

    6ab = (a + b)\textasciicircum{}2 \\

    \colorbox[HTML]{FFF9F9}{\strut But}\colorbox[HTML]{FFF5F5}{\strut  we}\colorbox[HTML]{FFE5E5}{\strut  also}\colorbox[HTML]{FFE8E8}{\strut  know}\colorbox[HTML]{FFD9D9}{\strut  that}\colorbox[HTML]{FFE8E8}{\strut  ab}\colorbox[HTML]{FFF8F8}{\strut  =}\colorbox[HTML]{FFFBFB}{\strut  }\colorbox[HTML]{FFE9E9}{\strut 54}\colorbox[HTML]{FFF1F1}{\strut ,}\colorbox[HTML]{FFE2E2}{\strut  so}\colorbox[HTML]{FFD3D3}{\strut  substitute}\colorbox[HTML]{FFE4E4}{\strut :} \\

    \colorbox[HTML]{FFFCFC}{\strut 6}\colorbox[HTML]{FFE8E8}{\strut *}\colorbox[HTML]{FFEFEF}{\strut 5}\colorbox[HTML]{FFEDED}{\strut 4}\colorbox[HTML]{FFF3F3}{\strut  =}\colorbox[HTML]{FFF7F7}{\strut  (}\colorbox[HTML]{FFF9F9}{\strut a}\colorbox[HTML]{FFE4E4}{\strut  +}\colorbox[HTML]{FFDFDF}{\strut  b}\colorbox[HTML]{FFCECE}{\strut )\textasciicircum{}}\colorbox[HTML]{FFD4D4}{\strut 2} \\

    \colorbox[HTML]{FFE0E0}{\strut 3}\colorbox[HTML]{FFDDDD}{\strut 2}\colorbox[HTML]{FFD6D6}{\strut 4}\colorbox[HTML]{FFDCDC}{\strut  =}\colorbox[HTML]{FFEAEA}{\strut  (}\colorbox[HTML]{FFF0F0}{\strut a +} b\colorbox[HTML]{FFF2F2}{\strut )\textasciicircum{}}\colorbox[HTML]{FFECEC}{\strut 2} \\

    \colorbox[HTML]{FFCECE}{\strut Therefore}\colorbox[HTML]{FFD6D6}{\strut ,} a\colorbox[HTML]{FFEAEA}{\strut  +}\colorbox[HTML]{FFF1F1}{\strut  b}\colorbox[HTML]{FFE0E0}{\strut  =}\colorbox[HTML]{FFBFBF}{\strut  sqrt}\colorbox[HTML]{FFEBEB}{\strut (}\colorbox[HTML]{FFF0F0}{\strut 3}\colorbox[HTML]{FFEAEA}{\strut 2}\colorbox[HTML]{FFE7E7}{\strut 4}\colorbox[HTML]{FFEBEB}{\strut )}\colorbox[HTML]{FFEDED}{\strut  = }\colorbox[HTML]{FFF2F2}{\strut 1}\colorbox[HTML]{FFF6F6}{\strut 8} \\

    \colorbox[HTML]{FFDBDB}{\strut So}\colorbox[HTML]{FFE8E8}{\strut  then} r\colorbox[HTML]{FFFCFC}{\strut  =} (a\colorbox[HTML]{FFE8E8}{\strut  +}\colorbox[HTML]{FFEDED}{\strut  b}\colorbox[HTML]{FFE4E4}{\strut )/}\colorbox[HTML]{FFEAEA}{\strut 6}\colorbox[HTML]{FFEBEB}{\strut  =}\colorbox[HTML]{FFEAEA}{\strut  }\colorbox[HTML]{FFD7D7}{\strut 1}\colorbox[HTML]{FFD6D6}{\strut 8}\colorbox[HTML]{FFCECE}{\strut  /}\colorbox[HTML]{FFEBEB}{\strut  }\colorbox[HTML]{FFEDED}{\strut 6}\colorbox[HTML]{FFEEEE}{\strut  =}\colorbox[HTML]{FFEDED}{\strut  }\colorbox[HTML]{FFE2E2}{\strut 3} \\

    \colorbox[HTML]{FFC9C9}{\strut Wait}\colorbox[HTML]{FFDBDB}{\strut ,}\colorbox[HTML]{FFD1D1}{\strut  so}\colorbox[HTML]{FFCECE}{\strut  the}\colorbox[HTML]{FFADAD}{\strut  radius}\colorbox[HTML]{FFDBDB}{\strut  is}\colorbox[HTML]{FFE0E0}{\strut  }\colorbox[HTML]{FFD9D9}{\strut 3}\colorbox[HTML]{FFC6C6}{\strut ?}\colorbox[HTML]{FFBCBC}{\strut  Let}\colorbox[HTML]{FFC8C8}{\strut  me}\colorbox[HTML]{FFAAAA}{\strut  check}\colorbox[HTML]{FFC9C9}{\strut  if}\colorbox[HTML]{FFCCCC}{\strut  that}\colorbox[HTML]{FFC7C7}{\strut  makes}\colorbox[HTML]{FFAFAF}{\strut  sense}\colorbox[HTML]{FFD2D2}{\strut .} \\

    \colorbox[HTML]{FFD3D3}{\strut Wait}\colorbox[HTML]{FFDDDD}{\strut ,}\colorbox[HTML]{FFD6D6}{\strut  so}\colorbox[HTML]{FFC9C9}{\strut  starting from}\colorbox[HTML]{FFDCDC}{\strut  the}\colorbox[HTML]{FFC3C3}{\strut  center}\colorbox[HTML]{FFEAEA}{\strut  being}\colorbox[HTML]{FFE4E4}{\strut  (}\colorbox[HTML]{FFEDED}{\strut r}\colorbox[HTML]{FFE7E7}{\strut ,}\colorbox[HTML]{FFE5E5}{\strut  r}\colorbox[HTML]{FFDEDE}{\strut )}\colorbox[HTML]{FFD5D5}{\strut  because}\colorbox[HTML]{FFE2E2}{\strut  it}\colorbox[HTML]{FFDEDE}{\strut 's}\colorbox[HTML]{FFC6C6}{\strut  distance}\colorbox[HTML]{FFF4F4}{\strut  r}\colorbox[HTML]{FFF7F7}{\strut  from}\\\colorbox[HTML]{FFEBEB}{\strut  both}\colorbox[HTML]{FFC7C7}{\strut  axes}\colorbox[HTML]{FFEFEF}{\strut ,}\colorbox[HTML]{FFFDFD}{\strut  and}\colorbox[HTML]{FFF2F2}{\strut  since} it\colorbox[HTML]{FFF4F4}{\strut 's}\colorbox[HTML]{FFEDED}{\strut  on}\colorbox[HTML]{FFDFDF}{\strut  AB}\colorbox[HTML]{FFEAEA}{\strut ,}\colorbox[HTML]{FFD4D4}{\strut  substit}\colorbox[HTML]{FFCCCC}{\strut uting}\colorbox[HTML]{FFCBCB}{\strut  into}\colorbox[HTML]{FFD1D1}{\strut  the}\colorbox[HTML]{FFB4B4}{\strut  equation}\colorbox[HTML]{FFD5D5}{\strut  gives}\colorbox[HTML]{FFE3E3}{\strut  r}\colorbox[HTML]{FFD7D7}{\strut  =}\colorbox[HTML]{FFD3D3}{\strut  (-}\colorbox[HTML]{FFD9D9}{\strut b}\colorbox[HTML]{FFCBCB}{\strut /a}\colorbox[HTML]{FFF5F5}{\strut )r}\colorbox[HTML]{FFE4E4}{\strut  +}\colorbox[HTML]{FFEBEB}{\strut  b}\colorbox[HTML]{FFDCDC}{\strut .}\\\colorbox[HTML]{FFCDCD}{\strut  Then}\colorbox[HTML]{FFA7A7}{\strut  solving}\colorbox[HTML]{FFD7D7}{\strut  that}\colorbox[HTML]{FFE3E3}{\strut  gives}\colorbox[HTML]{FFD5D5}{\strut  r}\colorbox[HTML]{FFC5C5}{\strut  =}\colorbox[HTML]{FFB0B0}{\strut  (}\colorbox[HTML]{FF9A9A}{\strut ab}\colorbox[HTML]{FFA2A2}{\strut )/(}\colorbox[HTML]{FF9B9B}{\strut a}\colorbox[HTML]{FFBBBB}{\strut  +}\colorbox[HTML]{FFA3A3}{\strut  b}\colorbox[HTML]{FFA5A5}{\strut ).}\colorbox[HTML]{FFC1C1}{\strut  Then}\colorbox[HTML]{FFA8A8}{\strut  according}\colorbox[HTML]{FFC5C5}{\strut  to}\colorbox[HTML]{FFBDBD}{\strut  the}\colorbox[HTML]{FFB9B9}{\strut  problem}\\\colorbox[HTML]{FFBABA}{\strut ,}\colorbox[HTML]{FFBBBB}{\strut  r}\colorbox[HTML]{FFB0B0}{\strut  is}\colorbox[HTML]{FFC5C5}{\strut  (}\colorbox[HTML]{FFDBDB}{\strut a}\colorbox[HTML]{FFCDCD}{\strut  +}\colorbox[HTML]{FFADAD}{\strut  b}\colorbox[HTML]{FF9F9F}{\strut )/}\colorbox[HTML]{FF9292}{\strut 6}\colorbox[HTML]{FF9898}{\strut .}\colorbox[HTML]{FF8888}{\strut  So}\colorbox[HTML]{FC6060}{\strut  equ}\colorbox[HTML]{FF8181}{\strut ate}\colorbox[HTML]{FF9595}{\strut  them}\colorbox[HTML]{FF6C6C}{\strut :} \\

    \colorbox[HTML]{FFA3A3}{\strut (ab}\colorbox[HTML]{FFAAAA}{\strut )/(}\colorbox[HTML]{FFC9C9}{\strut a}\colorbox[HTML]{FFCECE}{\strut  +}\colorbox[HTML]{FFBBBB}{\strut  b}\colorbox[HTML]{FFB0B0}{\strut )}\colorbox[HTML]{FF9898}{\strut  =}\colorbox[HTML]{FFC4C4}{\strut  (}\colorbox[HTML]{FFDCDC}{\strut a}\colorbox[HTML]{FFCBCB}{\strut  +}\colorbox[HTML]{FFACAC}{\strut  b}\colorbox[HTML]{FFA2A2}{\strut )/}\colorbox[HTML]{FF9393}{\strut 6} \\

    \colorbox[HTML]{FF6666}{\strut Cross}\colorbox[HTML]{FF6C6C}{\strut  multiply}\colorbox[HTML]{F85959}{\strut :} \\

    \colorbox[HTML]{EC4A4A}{\strut 6}\colorbox[HTML]{FF7979}{\strut ab}\colorbox[HTML]{F45555}{\strut  =}\colorbox[HTML]{FF6B6B}{\strut  (}\colorbox[HTML]{FF8B8B}{\strut a}\colorbox[HTML]{FF7878}{\strut  +}\colorbox[HTML]{FF6C6C}{\strut  b}\colorbox[HTML]{E84545}{\strut )\textasciicircum{}}\colorbox[HTML]{FD6161}{\strut 2} \\

    \colorbox[HTML]{EF4E4E}{\strut But}\colorbox[HTML]{E33D3D}{\strut  since}\colorbox[HTML]{E43F3F}{\strut  ab}\colorbox[HTML]{D12626}{\strut  =}\colorbox[HTML]{ED4B4B}{\strut  }\colorbox[HTML]{D12626}{\strut 5}\colorbox[HTML]{EF4E4E}{\strut 4}\colorbox[HTML]{D12626}{\strut ,}\colorbox[HTML]{DD3535}{\strut  }\colorbox[HTML]{B50000}{\strut 6}\colorbox[HTML]{BE0D0D}{\strut *}\colorbox[HTML]{FFD900}{\strut 5}
    \par}
\end{tcolorbox}

\vspace{1em}

\begin{tcolorbox}[
    enhanced, breakable,
    colback=white, colframe=black, boxrule=0.5pt,
    arc=2pt, left=4pt, right=4pt, top=4pt, bottom=4pt,
    fonttitle=\bfseries,
    width=\linewidth,
]
{\raggedright \ttfamily \footnotesize
    \textbf{Question:} \colorbox[HTML]{FFF7F7}{\strut <|begin of sentence|>}\colorbox[HTML]{FFF5F5}{\strut <|User|>}\colorbox[HTML]{FFFCFC}{\strut The}\colorbox[HTML]{FFF8F8}{\strut  solution} set of\colorbox[HTML]{FFF0F0}{\strut  \$\textbackslash\{\}}log\_\{\textbackslash\{\}frac\{1\}\{2\}\}\textbar{}x- \textbackslash\{\}frac\{\textbackslash\{\}pi\}\{3\}\textbar{} \textbackslash\{\}geq \textbackslash\{\}log\_\{\textbackslash\{\}frac\{1\}\{2\}\} \textbackslash\{\}frac\{\textbackslash\{\}pi\}\{2\}\$ is ( ) \\

    A: \$\textbackslash\{\}\{x\textbar{}-\textbackslash\{\}frac\{\textbackslash\{\}pi\}\{6\} \textbackslash\{\}leq x \textbackslash\{\}leq \textbackslash\{\}frac\{5\textbackslash\{\}pi\}\{6\}\textbackslash\{\}\}\$ \\

    B: \$\textbackslash\{\}\{x\textbar{}x \textbackslash\{\}leq -\textbackslash\{\}frac\{\textbackslash\{\}pi\}\{6\}, \textbackslash\{\}text\{ or \} x \textbackslash\{\}geq \textbackslash\{\}frac\{5\textbackslash\{\}pi\}\{6\}\textbackslash\{\}\}\$ \\

    C: \$\textbackslash\{\}\{x\textbar{}-\textbackslash\{\}frac\{\textbackslash\{\}pi\}\{6\} \textbackslash\{\}leq x \textbackslash\{\}leq \textbackslash\{\}frac\{5\textbackslash\{\}pi\}\{6\} \textbackslash\{\}text\{ and \} x \textbackslash\{\}neq \textbackslash\{\}frac\{\textbackslash\{\}pi\}\{3\}\textbackslash\{\}\}\$ \\

    D: \$\textbackslash\{\}\{x\textbar{}-\textbackslash\{\}frac\{5\textbackslash\{\}pi\}\{6\} \textbackslash\{\}leq x \textbackslash\{\}leq \textbackslash\{\}frac\{5\textbackslash\{\}pi\}\{6\} \textbackslash\{\}text\{ and \} x \textbackslash\{\}neq \textbackslash\{\}frac\{\textbackslash\{\}pi\}\{3\}\textbackslash\{\}\}\$<|Assistant|><think> \\
    \vspace{4pt} \hrule \vspace{4pt}
    \textbf{Response:} <think> \\
    Okay, so\colorbox[HTML]{FFF7F7}{\strut  I} need to solve this logarithmic inequality: log base 1/2 of \textbar{}x - $\pi$/3\textbar{} is greater than or equal to log base 1/2 of $\pi$/2. Then choose the correct solution set from the options A to D. Let me think step by step. \\

    First, I remember that logarithmic functions have certain properties, especially when dealing with inequalities. Since the base here is 1/2, which is between 0 and 1, the logarithmic function is decreasing. That means when I apply it to both sides of an inequality, the direction of the inequality will reverse. So, if log\_b(a) $\geq$ log\_b(c), then for 0 < b < 1, a $\leq$ c. Right? Let me confirm that. Yes, because as the input to a decreasing function increases, the output decreases. So, the original inequality log\_\{1/2\}(A) $\geq$ log\_\{1/2\}(B) implies A $\leq$ B. \\

    Therefore, applying this property here, the given inequality log\_\{1/2\}\textbar{}x - $\pi$/3\textbar{} $\geq$ log\_\{1/2\}($\pi$/2) implies that \textbar{}x - $\pi$/3\textbar{} $\leq$ $\pi$/2. But wait, also, we need to consider the domain of the logarithm. The argument of the logarithm must be positive. So, \textbar{}x - $\pi$/3\textbar{} > 0. Which means x $\neq$ $\pi$/3. Because absolute value is always non-negative, but logarithm is undefined at 0. So x cannot be $\pi$/3. \\

    So, putting this together, the solution set is the set of all x such that \textbar{}x - $\pi$/3\textbar{} $\leq$ $\pi$/2 and x $\neq$ $\pi$/3. Now, let's solve \textbar{}x - $\pi$/3\textbar{} $\leq$ $\pi$/2. The absolute value inequality \textbar{}A\colorbox[HTML]{FFFDFD}{\strut \textbar{}}\colorbox[HTML]{FFECEC}{\strut  $\leq$} B (\colorbox[HTML]{FFEEEE}{\strut where} B\colorbox[HTML]{FFF7F7}{\strut  >} 0) is equivalent to -B $\leq$ A $\leq$ B. So\colorbox[HTML]{FFF3F3}{\strut  here},\colorbox[HTML]{FFFCFC}{\strut  -}\colorbox[HTML]{FFF4F4}{\strut $\pi$}\colorbox[HTML]{FFFCFC}{\strut /}2 $\leq$ x - $\pi$/3 $\leq$ $\pi$/2. Let's solve\colorbox[HTML]{FFFCFC}{\strut  for} x. \\

    \colorbox[HTML]{FFFCFC}{\strut First}, add $\pi$/3 to all parts: \\

    -$\pi$/2 + $\pi$/3 $\leq$ x $\leq$ $\pi$/2 + $\pi$/3 \\

    \colorbox[HTML]{FFF2F2}{\strut Compute} -$\pi$/2 + $\pi$/3. To add these, find a common denominator. 6 is the least common multiple\colorbox[HTML]{FFFAFA}{\strut  of} 2 and 3. So: \\

    -$\pi$/2 = -3$\pi$/6 \\

    $\pi$/3 = 2$\pi$/6 \\

    So, -3$\pi$/6 + 2$\pi$/6 = (-3$\pi$ + 2$\pi$)/6 = -$\pi$/6 \\

    \colorbox[HTML]{FFFEFE}{\strut Similarly}, $\pi$/2 + $\pi$/3: \\

    $\pi$/2 = 3$\pi$/6 \\

    $\pi$/3 = 2$\pi$/6 \\

    3$\pi$/6 + 2$\pi$/6 = \colorbox[HTML]{FFF2F2}{\strut 5}\colorbox[HTML]{FFF3F3}{\strut $\pi$}\colorbox[HTML]{FFFDFD}{\strut /}\colorbox[HTML]{FFFEFE}{\strut 6} \\

    \colorbox[HTML]{FFEBEB}{\strut Therefore},\colorbox[HTML]{FFF9F9}{\strut  the}\colorbox[HTML]{FFE9E9}{\strut  inequality}\colorbox[HTML]{FFF6F6}{\strut  \textbar{}}x\colorbox[HTML]{FFF5F5}{\strut  -} $\pi$/3\textbar{}\colorbox[HTML]{FFF7F7}{\strut  $\leq$} $\pi$\colorbox[HTML]{FFFEFE}{\strut /}2\colorbox[HTML]{FFE2E2}{\strut  simpl}\colorbox[HTML]{FFF2F2}{\strut ifies} to -$\pi$/6 $\leq$ x $\leq$ 5$\pi$/6. But remember x\colorbox[HTML]{FFF4F4}{\strut  $\neq$} $\pi$/3 because \textbar{}x - $\pi$/3\textbar{} > 0. \\

    So, the\colorbox[HTML]{FFEEEE}{\strut  solution}\colorbox[HTML]{FFFEFE}{\strut  set} is all x\colorbox[HTML]{FFF9F9}{\strut  in}\colorbox[HTML]{FFF7F7}{\strut  [-}$\pi$/6, 5$\pi$/6] except x = $\pi$/3. Now,\colorbox[HTML]{FFFBFB}{\strut  looking}\colorbox[HTML]{FFFEFE}{\strut  at} the options: \\

    A: \{x \textbar{} -$\pi$/6 $\leq$ x $\leq$ 5$\pi$/6\} $\to$ Doesn't exclude $\pi$/3, so\colorbox[HTML]{FFFEFE}{\strut  incorrect}. \\

    B: \{x \textbar{} x $\leq$ -$\pi$/6 or x $\geq$ 5$\pi$/6\}\colorbox[HTML]{FFF4F4}{\strut  $\to$}\colorbox[HTML]{FFEDED}{\strut  This} is the\colorbox[HTML]{FFF9F9}{\strut  complement}\colorbox[HTML]{FFFCFC}{\strut  of} the\colorbox[HTML]{FFFDFD}{\strut  interval}, which would be if the inequality was\colorbox[HTML]{FFF3F3}{\strut  reversed}. But\colorbox[HTML]{FFF9F9}{\strut  since}\colorbox[HTML]{FFFAFA}{\strut  our}\colorbox[HTML]{FFF3F3}{\strut  inequality} was \textbar{}x - $\pi$/3\textbar{} $\leq$ $\pi$/2, this is\colorbox[HTML]{FFFFFF}{\strut  the}\colorbox[HTML]{FFE2E2}{\strut  outside}\colorbox[HTML]{FFF1F1}{\strut ,}\colorbox[HTML]{FFE8E8}{\strut  which}\colorbox[HTML]{FFF6F6}{\strut  is}\colorbox[HTML]{FFE1E1}{\strut  wrong}\colorbox[HTML]{FFE7E7}{\strut .} \\

    \colorbox[HTML]{FFD9D9}{\strut C}\colorbox[HTML]{FFE9E9}{\strut :}\colorbox[HTML]{FFE0E0}{\strut  \{}\colorbox[HTML]{FFE2E2}{\strut x}\colorbox[HTML]{FFFAFA}{\strut  \textbar{}}\colorbox[HTML]{FFEDED}{\strut  -}\colorbox[HTML]{FFF0F0}{\strut $\pi$}\colorbox[HTML]{FFEEEE}{\strut /}\colorbox[HTML]{FFEFEF}{\strut 6}\colorbox[HTML]{FFE9E9}{\strut  $\leq$}\colorbox[HTML]{FFF3F3}{\strut  x}\colorbox[HTML]{FFE9E9}{\strut  $\leq$}\colorbox[HTML]{FFE0E0}{\strut  }\colorbox[HTML]{FFDEDE}{\strut 5}\colorbox[HTML]{FFDADA}{\strut $\pi$}\colorbox[HTML]{FFD8D8}{\strut /}\colorbox[HTML]{FFDEDE}{\strut 6}\colorbox[HTML]{FFD8D8}{\strut  and}\colorbox[HTML]{FFDBDB}{\strut  x}\colorbox[HTML]{FFCECE}{\strut  $\neq$}\colorbox[HTML]{FFCFCF}{\strut  $\pi$}\colorbox[HTML]{FFDEDE}{\strut /}\colorbox[HTML]{FFE9E9}{\strut 3}\colorbox[HTML]{FFE5E5}{\strut \}}\colorbox[HTML]{FFDCDC}{\strut  $\to$}\colorbox[HTML]{FFE7E7}{\strut  This}\colorbox[HTML]{FFD1D1}{\strut  matches}\colorbox[HTML]{FFEFEF}{\strut  our}\\\colorbox[HTML]{FFD9D9}{\strut  solution}\colorbox[HTML]{FFEFEF}{\strut .}\colorbox[HTML]{FFE2E2}{\strut  Includes}\colorbox[HTML]{FFEBEB}{\strut  the}\colorbox[HTML]{FFE0E0}{\strut  interval}\colorbox[HTML]{FFE2E2}{\strut  and}\colorbox[HTML]{FFCECE}{\strut  excludes}\colorbox[HTML]{FFE8E8}{\strut  $\pi$}\colorbox[HTML]{FFE3E3}{\strut /}\colorbox[HTML]{FFF1F1}{\strut 3}\colorbox[HTML]{FFD9D9}{\strut .} \\

    \colorbox[HTML]{FFCCCC}{\strut D}\colorbox[HTML]{FFE9E9}{\strut :}\colorbox[HTML]{FFDEDE}{\strut  \{}\colorbox[HTML]{FFF5F5}{\strut x}\colorbox[HTML]{FFF8F8}{\strut  \textbar{}}\colorbox[HTML]{FFEAEA}{\strut  -}\colorbox[HTML]{FFEFEF}{\strut 5}\colorbox[HTML]{FFFEFE}{\strut $\pi$}\colorbox[HTML]{FFF9F9}{\strut /}\colorbox[HTML]{FFFEFE}{\strut 6}\colorbox[HTML]{FFF8F8}{\strut  $\leq$}\colorbox[HTML]{FFF6F6}{\strut  x}\colorbox[HTML]{FFEEEE}{\strut  $\leq$}\colorbox[HTML]{FFD3D3}{\strut  }\colorbox[HTML]{FFE7E7}{\strut 5}\colorbox[HTML]{FFF1F1}{\strut $\pi$}\colorbox[HTML]{FFE1E1}{\strut /}\colorbox[HTML]{FFE0E0}{\strut 6}\colorbox[HTML]{FFCFCF}{\strut  and}\colorbox[HTML]{FFD1D1}{\strut  x}\colorbox[HTML]{FFCFCF}{\strut  $\neq$}\colorbox[HTML]{FFE1E1}{\strut  $\pi$}\colorbox[HTML]{FFE1E1}{\strut /}\colorbox[HTML]{FFEBEB}{\strut 3}\colorbox[HTML]{FFE0E0}{\strut \}}\colorbox[HTML]{FFD8D8}{\strut  $\to$}\colorbox[HTML]{FFCDCD}{\strut  The}\colorbox[HTML]{FFDEDE}{\strut  interval}\colorbox[HTML]{FFE1E1}{\strut  here}\\\colorbox[HTML]{FFE6E6}{\strut  is}\colorbox[HTML]{FFE0E0}{\strut  from}\colorbox[HTML]{FFDCDC}{\strut  -}\colorbox[HTML]{FFD4D4}{\strut 5}\colorbox[HTML]{FFE4E4}{\strut $\pi$}\colorbox[HTML]{FFD9D9}{\strut /}\colorbox[HTML]{FFDCDC}{\strut 6}\colorbox[HTML]{FFD5D5}{\strut  to }\colorbox[HTML]{FFBDBD}{\strut 5}\colorbox[HTML]{FFDBDB}{\strut $\pi$/}\colorbox[HTML]{FFDBDB}{\strut 6}\colorbox[HTML]{FFDDDD}{\strut ,}\colorbox[HTML]{FFD6D6}{\strut  which}\colorbox[HTML]{FFE0E0}{\strut  is}\colorbox[HTML]{FFD2D2}{\strut  different}\colorbox[HTML]{FFECEC}{\strut .}\colorbox[HTML]{FFE0E0}{\strut  Let}\colorbox[HTML]{FFE8E8}{\strut  me}\colorbox[HTML]{FFDADA}{\strut  check}\colorbox[HTML]{FFE3E3}{\strut  if}\colorbox[HTML]{FFF4F4}{\strut  I}\colorbox[HTML]{FFE6E6}{\strut  made}\colorbox[HTML]{FFEAEA}{\strut  a}\\\colorbox[HTML]{FFDBDB}{\strut  mistake}\colorbox[HTML]{FFE3E3}{\strut  in}\colorbox[HTML]{FFB5B5}{\strut  solving}\colorbox[HTML]{FFDBDB}{\strut  the}\colorbox[HTML]{FFDADA}{\strut  inequality}\colorbox[HTML]{FFD4D4}{\strut .} \\

    \colorbox[HTML]{FFD6D6}{\strut Wait}\colorbox[HTML]{FFE5E5}{\strut ,}\colorbox[HTML]{FFDADA}{\strut  when solving}\colorbox[HTML]{FFEEEE}{\strut  \textbar{}}\colorbox[HTML]{FFFBFB}{\strut x}\colorbox[HTML]{FFF3F3}{\strut  -}\colorbox[HTML]{FFF3F3}{\strut  $\pi$}\colorbox[HTML]{FFF4F4}{\strut /}\colorbox[HTML]{FFF3F3}{\strut 3}\colorbox[HTML]{FFFFFF}{\strut \textbar{}}\colorbox[HTML]{FFFAFA}{\strut  $\leq$}\colorbox[HTML]{FFF3F3}{\strut  $\pi$}\colorbox[HTML]{FFE9E9}{\strut /}\colorbox[HTML]{FFE6E6}{\strut 2}\colorbox[HTML]{FFDFDF}{\strut ,}\colorbox[HTML]{FFCACA}{\strut  adding}\colorbox[HTML]{FFE9E9}{\strut  $\pi$}\colorbox[HTML]{FFE6E6}{\strut /}\colorbox[HTML]{FFEDED}{\strut 3}\colorbox[HTML]{FFEEEE}{\strut  to}\colorbox[HTML]{FFE0E0}{\strut  -}\colorbox[HTML]{FFDFDF}{\strut $\pi$}\colorbox[HTML]{FFF1F1}{\strut /}\colorbox[HTML]{FFF1F1}{\strut 2}\colorbox[HTML]{FFE5E5}{\strut  gives}\\\colorbox[HTML]{FFE8E8}{\strut  -}\colorbox[HTML]{FFE9E9}{\strut $\pi$}\colorbox[HTML]{FFEDED}{\strut /}\colorbox[HTML]{FFE0E0}{\strut 6}\colorbox[HTML]{FFE1E1}{\strut ,}\colorbox[HTML]{FFDEDE}{\strut  and}\colorbox[HTML]{FFE2E2}{\strut  adding}\colorbox[HTML]{FFD1D1}{\strut  $\pi$}\colorbox[HTML]{FFC5C5}{\strut /}\colorbox[HTML]{FFB8B8}{\strut 3}\colorbox[HTML]{FFD3D3}{\strut  to}\colorbox[HTML]{FFCECE}{\strut  $\pi$}\colorbox[HTML]{FFD4D4}{\strut /}\colorbox[HTML]{FFD0D0}{\strut 2}\colorbox[HTML]{FFD8D8}{\strut  gives}\colorbox[HTML]{FFCACA}{\strut  }\colorbox[HTML]{FFB5B5}{\strut 5}\colorbox[HTML]{FFDCDC}{\strut $\pi$}\colorbox[HTML]{FFE4E4}{\strut /}\colorbox[HTML]{FFDCDC}{\strut 6}\colorbox[HTML]{FFDBDB}{\strut .}\colorbox[HTML]{FFCCCC}{\strut  That}\colorbox[HTML]{FFD9D9}{\strut 's}\colorbox[HTML]{FFD0D0}{\strut  correct}\colorbox[HTML]{FFD5D5}{\strut .}\\\colorbox[HTML]{FFC5C5}{\strut  So}\colorbox[HTML]{FFB4B4}{\strut  D}\colorbox[HTML]{FFCCCC}{\strut  is}\colorbox[HTML]{FFBDBD}{\strut  incorrect}\colorbox[HTML]{FFB6B6}{\strut  because}\colorbox[HTML]{FFC7C7}{\strut  the}\colorbox[HTML]{FFB9B9}{\strut  interval}\colorbox[HTML]{FFD2D2}{\strut  is}\colorbox[HTML]{FFC4C4}{\strut  from}\colorbox[HTML]{FFBCBC}{\strut  -}\colorbox[HTML]{FFA2A2}{\strut 5}\colorbox[HTML]{FFD7D7}{\strut $\pi$}\colorbox[HTML]{FFE2E2}{\strut /}\colorbox[HTML]{FFDCDC}{\strut 6}\colorbox[HTML]{FFCECE}{\strut ,}\colorbox[HTML]{FFCCCC}{\strut  which}\colorbox[HTML]{FFD4D4}{\strut  is}\colorbox[HTML]{FFD9D9}{\strut  more}\\\colorbox[HTML]{FF9B9B}{\strut  negative}\colorbox[HTML]{FFBABA}{\strut .}\colorbox[HTML]{FFC1C1}{\strut  So}\colorbox[HTML]{FFA2A2}{\strut  that}\colorbox[HTML]{FFB1B1}{\strut 's}\colorbox[HTML]{FFA5A5}{\strut  not}\colorbox[HTML]{FF9999}{\strut  matching}\colorbox[HTML]{FFBABA}{\strut .}\colorbox[HTML]{FFA4A4}{\strut  So}\colorbox[HTML]{FF8888}{\strut  the}\colorbox[HTML]{FF7474}{\strut  answer}\colorbox[HTML]{FF9797}{\strut  must}\colorbox[HTML]{FFA8A8}{\strut  be}\colorbox[HTML]{FF7474}{\strut  C}\colorbox[HTML]{FF7D7D}{\strut .} \\

    \colorbox[HTML]{FF9B9B}{\strut But}\colorbox[HTML]{FF6F6F}{\strut  let}\colorbox[HTML]{FFABAB}{\strut  me}\colorbox[HTML]{FFB0B0}{\strut  just}\colorbox[HTML]{FF7B7B}{\strut  verify}\colorbox[HTML]{FF8C8C}{\strut  once}\colorbox[HTML]{FF8989}{\strut  more}\colorbox[HTML]{FF6767}{\strut .}\colorbox[HTML]{FF7D7D}{\strut  The}\colorbox[HTML]{F04F4F}{\strut  original}\colorbox[HTML]{FF6666}{\strut  inequality}\colorbox[HTML]{FF6A6A}{\strut  is}\colorbox[HTML]{B50000}{\strut  log}\colorbox[HTML]{C21212}{\strut \_\{}\colorbox[HTML]{D32828}{\strut 1}\colorbox[HTML]{FFD900}{\strut /}
    \par}
\end{tcolorbox}

\subsection{Layer~26 Indexer Output Distribution}

\begin{tcolorbox}[
    enhanced, breakable,
    colback=white, colframe=black, boxrule=0.5pt,
    arc=2pt, left=4pt, right=4pt, top=4pt, bottom=4pt,
    fonttitle=\bfseries,
    width=\linewidth,
]
{\raggedright \ttfamily \footnotesize
    \textbf{Question:} \colorbox[HTML]{B80505}{\strut <|begin of sentence|>}\colorbox[HTML]{B50000}{\strut <|User|>}\colorbox[HTML]{FFECEC}{\strut The} center of a circle touching the legs \$A C\$ and \$B C\$ of a right triangle \$A B C\$ lies on\colorbox[HTML]{FFFCFC}{\strut  the} hypotenuse \$A B\$. Find the radius of the circle, if it is six times smaller than the sum of the legs, and the area of triangle \$A B C\$ is 27.\colorbox[HTML]{FFDBDB}{\strut <|Assistant|>}\colorbox[HTML]{FFCACA}{\strut <think>} \\
    
    \vspace{4pt} \hrule \vspace{4pt}
    \textbf{Response:} \colorbox[HTML]{FFDDDD}{\strut <think>} \\
    Okay\colorbox[HTML]{FFCACA}{\strut ,} so I have this problem about a right triangle ABC where the circle touches the legs AC and BC, and its center is on the hypotenuse AB\colorbox[HTML]{FFFDFD}{\strut .} I need to find\colorbox[HTML]{FFF9F9}{\strut  the} radius of the circle\colorbox[HTML]{FFF4F4}{\strut ,} which is six times smaller than\colorbox[HTML]{FFF5F5}{\strut  the} sum of the legs\colorbox[HTML]{FFEBEB}{\strut .} Also, the area of the triangle is 27. Hmm\colorbox[HTML]{FFF6F6}{\strut ,} let me break this down step\colorbox[HTML]{FFDFDF}{\strut  by} step\colorbox[HTML]{FFE9E9}{\strut .} \\

    First, let me visualize the triangle. Since it's a right triangle, let's assume angle C is the right angle. So, legs AC and BC form the right angle, and AB is the hypotenuse. The circle touches both legs AC and BC, so it must be tangent to both. The center of the circle lies on AB\colorbox[HTML]{FFFAFA}{\strut .} That makes me think it's an incircle or maybe an excircle\colorbox[HTML]{FFEFEF}{\strut ,} but since it's touching both legs and the center is on the hypotenuse, it's probably not the standard incircle because the incenter of a right triangle is located inside the triangle\colorbox[HTML]{FFFBFB}{\strut ,} not on the hypotenuse\colorbox[HTML]{FFF5F5}{\strut .} Wait, but the problem says the center lies on the hypotenuse. So this must be a different circle, not the incircle\colorbox[HTML]{FFE9E9}{\strut .} \\

    Let me recall that if a circle is tangent to both legs of a right triangle and its center is on the hypotenuse, there might be a specific formula or relation here\colorbox[HTML]{FFF3F3}{\strut .} Let me denote the legs as AC = b and BC = a, and hypotenuse AB = c\colorbox[HTML]{FFEDED}{\strut .}\colorbox[HTML]{FFFAFA}{\strut  The} area is (a*b)/2 = 27\colorbox[HTML]{FFF5F5}{\strut ,} so a*b = 54\colorbox[HTML]{FFFFFF}{\strut .} Also, the radius r is six times smaller than the sum of the legs\colorbox[HTML]{FFF3F3}{\strut ,} so r = (a + b)/6\colorbox[HTML]{FFF7F7}{\strut .} My goal is to find r\colorbox[HTML]{FFDDDD}{\strut .} \\

    Wait\colorbox[HTML]{FFF1F1}{\strut ,} so I need to relate r to a and b\colorbox[HTML]{FFF4F4}{\strut .} Let me think about the coordinates. Maybe coordinate geometry can help here\colorbox[HTML]{FFF5F5}{\strut .} Let me place point C at the origin (0,0\colorbox[HTML]{FFFFFF}{\strut ),} point B at (a,0), point A at (0,b). Then hypotenuse AB goes from (0,b) to (a,0). The center of the circle lies on AB and is equidistant from both legs AC and BC\colorbox[HTML]{FFE8E8}{\strut .} Since the circle is tangent to both legs, which are the y-axis and x-axis respectively\colorbox[HTML]{FFFDFD}{\strut ,} the distance from the center to each axis must be equal to the radius\colorbox[HTML]{FFD7D7}{\strut .} Wait\colorbox[HTML]{FFD4D4}{\strut ,} if the center is (r, r\colorbox[HTML]{FFEEEE}{\strut )} for\colorbox[HTML]{FFF4F4}{\strut  the} incircle\colorbox[HTML]{FFEFEF}{\strut ,} but in this case, the center is on AB\colorbox[HTML]{FFD5D5}{\strut .} So\colorbox[HTML]{FFFBFB}{\strut  the} center (h,k) must satisfy h = k = r\colorbox[HTML]{FFC8C8}{\strut ?} No\colorbox[HTML]{FFEAEA}{\strut ,} because\colorbox[HTML]{FFFEFE}{\strut  the} center is on AB\colorbox[HTML]{FFD7D7}{\strut .} Wait\colorbox[HTML]{FFEDED}{\strut ,} maybe the coordinates of the center can be parametrized\colorbox[HTML]{FFF4F4}{\strut  along} the hypotenuse\colorbox[HTML]{FFD2D2}{\strut .} \\

    Let\colorbox[HTML]{FFF9F9}{\strut  me} parametrize AB\colorbox[HTML]{FFECEC}{\strut .}\colorbox[HTML]{FFFEFE}{\strut  The} hypotenuse AB can be represented by the line from (0, b\colorbox[HTML]{FFF9F9}{\strut )} to (a, 0\colorbox[HTML]{FFC8C8}{\strut ).}\colorbox[HTML]{FFECEC}{\strut  The} equation\colorbox[HTML]{FFF7F7}{\strut  of} AB\colorbox[HTML]{FFF4F4}{\strut  is} y\colorbox[HTML]{FFE8E8}{\strut  =}\colorbox[HTML]{FFEBEB}{\strut  (-}b/a\colorbox[HTML]{FFE6E6}{\strut )x}\colorbox[HTML]{FFDDDD}{\strut  +}\colorbox[HTML]{FFFDFD}{\strut  b}\colorbox[HTML]{FFBABA}{\strut .} Any point on AB can be written as (x\colorbox[HTML]{FFEFEF}{\strut ,}\colorbox[HTML]{FFF1F1}{\strut  (-}b/a)x\colorbox[HTML]{FFFEFE}{\strut  +} b\colorbox[HTML]{FFD7D7}{\strut ).} The center of the circle is at some point (x, (-b/a)x + b\colorbox[HTML]{FFF5F5}{\strut )}\colorbox[HTML]{FFFDFD}{\strut  on} AB\colorbox[HTML]{FFF8F8}{\strut ,} and this point must be at a distance\colorbox[HTML]{FFD5D5}{\strut  r} from both legs AC and BC\colorbox[HTML]{FFD8D8}{\strut .}\colorbox[HTML]{FFF3F3}{\strut  The} legs AC and BC are the y-axis\colorbox[HTML]{FFF5F5}{\strut  and} x-axis respectively\colorbox[HTML]{FFDFDF}{\strut ,} so the distance from the center to AC (y-axis) is x\colorbox[HTML]{FFDFDF}{\strut ,}\colorbox[HTML]{FFE8E8}{\strut  and}\colorbox[HTML]{FFDFDF}{\strut  the} distance to BC (x-axis) is y\colorbox[HTML]{FFD9D9}{\strut .} Since the circle is tangent to both legs\colorbox[HTML]{FFD4D4}{\strut ,} these distances must be equal to\colorbox[HTML]{FFEDED}{\strut  the} radius\colorbox[HTML]{FFDBDB}{\strut .} So x\colorbox[HTML]{FFF4F4}{\strut  =}\colorbox[HTML]{FFFCFC}{\strut  r}\colorbox[HTML]{FFDCDC}{\strut  and} y =\colorbox[HTML]{FFFDFD}{\strut  r}\colorbox[HTML]{FFC7C7}{\strut .} Therefore\colorbox[HTML]{FFF3F3}{\strut ,}\colorbox[HTML]{FFF8F8}{\strut  the} center\colorbox[HTML]{FFFCFC}{\strut  is}\colorbox[HTML]{FFE9E9}{\strut  (}r\colorbox[HTML]{FFEAEA}{\strut ,} r\colorbox[HTML]{FFC0C0}{\strut ),}\colorbox[HTML]{FFD3D3}{\strut  but}\colorbox[HTML]{FFFDFD}{\strut  also} lies on AB\colorbox[HTML]{FFC8C8}{\strut .}\colorbox[HTML]{FFF4F4}{\strut  Wait}\colorbox[HTML]{FFCCCC}{\strut ,}\colorbox[HTML]{FFEEEE}{\strut  but}\colorbox[HTML]{FFFCFC}{\strut  if}\colorbox[HTML]{FFE5E5}{\strut  the} center is (r, r\colorbox[HTML]{FFD6D6}{\strut ),}\colorbox[HTML]{FFF1F1}{\strut  then} substit\colorbox[HTML]{FFF7F7}{\strut uting} into\colorbox[HTML]{FFF0F0}{\strut  the} equation of AB gives r =\colorbox[HTML]{FFFEFE}{\strut  (-}b/a)r +\colorbox[HTML]{FFFDFD}{\strut  b}\colorbox[HTML]{FFA4A4}{\strut .} Let\colorbox[HTML]{FFD6D6}{\strut  me} check that\colorbox[HTML]{FFAAAA}{\strut .} \\

    \colorbox[HTML]{FFD5D5}{\strut The} equation of AB is y = (-b/a)x + b\colorbox[HTML]{FFBEBE}{\strut .} So if the center is at (r, r\colorbox[HTML]{FFDBDB}{\strut ),}\colorbox[HTML]{FFE0E0}{\strut  then} substituting x = r into\colorbox[HTML]{FFF1F1}{\strut  the} equation gives y = (-b/a)r + b\colorbox[HTML]{FFDDDD}{\strut .} But\colorbox[HTML]{FFE8E8}{\strut  the} y-coordinate of the center is also r\colorbox[HTML]{FFE0E0}{\strut ,} so we have\colorbox[HTML]{FFE8E8}{\strut :} \\

    r = (-b/a)r +\colorbox[HTML]{FFFBFB}{\strut  b} \\

    Let\colorbox[HTML]{FFF9F9}{\strut  me} solve\colorbox[HTML]{FFF4F4}{\strut  for} r\colorbox[HTML]{FFCDCD}{\strut :} \\

    r + (b/a)r\colorbox[HTML]{FFF7F7}{\strut  =} b \\

    r(1 + b/a\colorbox[HTML]{FFFEFE}{\strut )} = b \\

    r\colorbox[HTML]{FFF4F4}{\strut  =} b / (1 + b/a\colorbox[HTML]{FFD1D1}{\strut )} = (\colorbox[HTML]{FFFDFD}{\strut ab}\colorbox[HTML]{FFFEFE}{\strut )}\colorbox[HTML]{FFFDFD}{\strut  /} (a + b\colorbox[HTML]{FF8B8B}{\strut )} \\

    \colorbox[HTML]{FFFCFC}{\strut But} from the problem statement\colorbox[HTML]{FFFAFA}{\strut ,} the radius r is (a + b)/6\colorbox[HTML]{FFD4D4}{\strut .} So according to this\colorbox[HTML]{FFF4F4}{\strut ,} (ab)/(a + b\colorbox[HTML]{FFF4F4}{\strut )} = (a + b)/6 \\

    Multiply both sides by (a + b\colorbox[HTML]{FFE7E7}{\strut ):} \\

    ab\colorbox[HTML]{FFFDFD}{\strut  =} (a + b)\textasciicircum{}\colorbox[HTML]{FFF8F8}{\strut 2} / 6 \\

    Multiply both sides by 6: \\

    6ab\colorbox[HTML]{FFF0F0}{\strut  =} (a + b)\textasciicircum{}\colorbox[HTML]{FFE6E6}{\strut 2} \\

    \colorbox[HTML]{FFEDED}{\strut But}\colorbox[HTML]{FFE9E9}{\strut  we} also know\colorbox[HTML]{FFF7F7}{\strut  that} ab = 54\colorbox[HTML]{FFBBBB}{\strut ,}\colorbox[HTML]{FFE7E7}{\strut  so}\colorbox[HTML]{FFF9F9}{\strut  substitute}\colorbox[HTML]{FFFBFB}{\strut :} \\

    6\colorbox[HTML]{FFEEEE}{\strut *}54\colorbox[HTML]{FFEFEF}{\strut  =} (a\colorbox[HTML]{FFFFFF}{\strut  +} b\colorbox[HTML]{FFE2E2}{\strut )\textasciicircum{}}\colorbox[HTML]{FFE6E6}{\strut 2} \\

    32\colorbox[HTML]{FFFBFB}{\strut 4}\colorbox[HTML]{FFE5E5}{\strut  =}\colorbox[HTML]{FFF1F1}{\strut  (}a + b)\textasciicircum{}\colorbox[HTML]{FFE3E3}{\strut 2} \\

    Therefore\colorbox[HTML]{FFD5D5}{\strut ,} a\colorbox[HTML]{FFF5F5}{\strut  +} b\colorbox[HTML]{FFE1E1}{\strut  =}\colorbox[HTML]{FFEAEA}{\strut  sqrt}\colorbox[HTML]{FFF8F8}{\strut (}32\colorbox[HTML]{FFF7F7}{\strut 4}\colorbox[HTML]{FFDFDF}{\strut )}\colorbox[HTML]{FFE2E2}{\strut  =}\colorbox[HTML]{FFF1F1}{\strut  }18 \\

    \colorbox[HTML]{FFF5F5}{\strut So}\colorbox[HTML]{FFF4F4}{\strut  then}\colorbox[HTML]{FFF8F8}{\strut  r} = (a + b)/6\colorbox[HTML]{FFF6F6}{\strut  =} 18\colorbox[HTML]{FFFAFA}{\strut  /} 6\colorbox[HTML]{FFFDFD}{\strut  =} \colorbox[HTML]{FFFDFD}{\strut 3} \\

    \colorbox[HTML]{FFF5F5}{\strut Wait}\colorbox[HTML]{FFC9C9}{\strut ,} so\colorbox[HTML]{FFE0E0}{\strut  the} radius\colorbox[HTML]{FFEBEB}{\strut  is} 3\colorbox[HTML]{FFC7C7}{\strut ?} Let\colorbox[HTML]{FFDCDC}{\strut  me} check if that makes sense\colorbox[HTML]{FFB1B1}{\strut .} \\

    \colorbox[HTML]{FFE9E9}{\strut Wait}\colorbox[HTML]{FFBFBF}{\strut ,}\colorbox[HTML]{FFEFEF}{\strut  so} starting from the center being (r, r\colorbox[HTML]{FFF3F3}{\strut )}\colorbox[HTML]{FFF4F4}{\strut  because}it's distance r from both axes\colorbox[HTML]{FFCFCF}{\strut ,}\colorbox[HTML]{FFEBEB}{\strut  and} since it's on AB\colorbox[HTML]{FFD5D5}{\strut ,} substit\colorbox[HTML]{FFF2F2}{\strut uting} into the equation gives r = (-b/a)r + b\colorbox[HTML]{FFA4A4}{\strut .}\colorbox[HTML]{FFF3F3}{\strut  Then} solving\colorbox[HTML]{FFFFFF}{\strut  that} gives r\colorbox[HTML]{FFF6F6}{\strut  = (}ab\colorbox[HTML]{FFE6E6}{\strut )/(}a + b\colorbox[HTML]{FF9696}{\strut ).}\colorbox[HTML]{FFF3F3}{\strut  Then}\colorbox[HTML]{FFFAFA}{\strut  according}\colorbox[HTML]{FFFDFD}{\strut  to}\colorbox[HTML]{FFE3E3}{\strut  the} problem\colorbox[HTML]{FFE1E1}{\strut ,} r\colorbox[HTML]{FFE1E1}{\strut  is} (a + b)/6\colorbox[HTML]{FF9696}{\strut .}\colorbox[HTML]{FFDADA}{\strut  So} equ\colorbox[HTML]{FFF3F3}{\strut ate}\colorbox[HTML]{FFF5F5}{\strut  them}\colorbox[HTML]{FFB1B1}{\strut :} \\

    \colorbox[HTML]{FFE8E8}{\strut (ab}\colorbox[HTML]{FFF5F5}{\strut )/(}a + b\colorbox[HTML]{FFB8B8}{\strut )}\colorbox[HTML]{FFD1D1}{\strut  =} (a +\colorbox[HTML]{FFF5F5}{\strut  b}\colorbox[HTML]{FFEEEE}{\strut )/}\colorbox[HTML]{FFD2D2}{\strut 6} \\

    \colorbox[HTML]{FFEFEF}{\strut Cross}\colorbox[HTML]{FFCDCD}{\strut  multiply}\colorbox[HTML]{FFBDBD}{\strut :} \\

    \colorbox[HTML]{FFE9E9}{\strut 6}\colorbox[HTML]{FFB8B8}{\strut ab}\colorbox[HTML]{FFA3A3}{\strut  =}\colorbox[HTML]{FFD6D6}{\strut  (}\colorbox[HTML]{FFE9E9}{\strut a}\colorbox[HTML]{FFE7E7}{\strut  +}\colorbox[HTML]{FFF1F1}{\strut  b}\colorbox[HTML]{FFC4C4}{\strut )\textasciicircum{}}\colorbox[HTML]{FFBCBC}{\strut 2} \\

    \colorbox[HTML]{FFB1B1}{\strut But}\colorbox[HTML]{FFC4C4}{\strut  since}\colorbox[HTML]{FFEDED}{\strut  ab}\colorbox[HTML]{FFA2A2}{\strut  =}\colorbox[HTML]{FF9999}{\strut  }\colorbox[HTML]{FFDADA}{\strut 5}\colorbox[HTML]{FFD3D3}{\strut 4}\colorbox[HTML]{FF8989}{\strut ,}\colorbox[HTML]{FFB2B2}{\strut  }\colorbox[HTML]{FFE9E9}{\strut 6}\colorbox[HTML]{F45454}{\strut *}\colorbox[HTML]{FFD900}{\strut 5}
    \par}
\end{tcolorbox}

\vspace{1em}

\begin{tcolorbox}[
    enhanced, breakable,
    colback=white, colframe=black, boxrule=0.5pt,
    arc=2pt, left=4pt, right=4pt, top=4pt, bottom=4pt,
    fonttitle=\bfseries,
    width=\linewidth,
]
{\raggedright \ttfamily \footnotesize
    \textbf{Question:} \colorbox[HTML]{B90606}{\strut <|begin of sentence|>}\colorbox[HTML]{B50000}{\strut <|User|>}\colorbox[HTML]{FFEFEF}{\strut The} solution set of \$\textbackslash\{\}log\_\{\colorbox[HTML]{FFF5F5}{\strut \textbackslash\{\}}frac\colorbox[HTML]{FFF9F9}{\strut \{}1\}\{2\}\}\textbar{}x- \textbackslash\{\}\colorbox[HTML]{FFFFFF}{\strut frac}\colorbox[HTML]{FFF4F4}{\strut \{\textbackslash\{\}}pi\}\{\colorbox[HTML]{FFFAFA}{\strut 3}\}\textbar{} \textbackslash\{\}geq \textbackslash\{\}log\_\{\colorbox[HTML]{FFF9F9}{\strut \textbackslash\{\}}\colorbox[HTML]{FFFEFE}{\strut frac}\{1\}\{2\}\} \textbackslash\{\}frac\{\textbackslash\{\}pi\}\{2\}\$ is (\colorbox[HTML]{FFECEC}{\strut  }\colorbox[HTML]{FFF3F3}{\strut  }\colorbox[HTML]{FFDBDB}{\strut )} \\

    \colorbox[HTML]{FFFAFA}{\strut A}: \$\textbackslash\{\}\{x\textbar{}\colorbox[HTML]{FFF5F5}{\strut -\textbackslash\{\}}\colorbox[HTML]{FFFDFD}{\strut frac}\colorbox[HTML]{FFF8F8}{\strut \{\textbackslash\{\}}pi\}\{6\} \textbackslash\{\}leq x \textbackslash\{\}leq \textbackslash\{\}frac\{5\textbackslash\{\}pi\}\{6\colorbox[HTML]{FFF5F5}{\strut \}\textbackslash\{\}}\colorbox[HTML]{FFF4F4}{\strut \}\$} \\

    \colorbox[HTML]{FFFDFD}{\strut B}: \$\textbackslash\{\}\{x\textbar{}x\colorbox[HTML]{FFFEFE}{\strut  \textbackslash\{\}}leq -\textbackslash\{\}frac\colorbox[HTML]{FFFDFD}{\strut \{\textbackslash\{\}}pi\}\{6\},\colorbox[HTML]{FFFEFE}{\strut  \textbackslash\{\}}text\{ or \} x \textbackslash\{\}geq \textbackslash\{\}frac\{5\textbackslash\{\}pi\}\{6\colorbox[HTML]{FFFBFB}{\strut \}\textbackslash\{\}}\colorbox[HTML]{FFFDFD}{\strut \}\$} \\

    C: \$\textbackslash\{\}\{x\textbar{}-\textbackslash\{\}frac\{\textbackslash\{\}pi\}\{6\} \textbackslash\{\}leq x \textbackslash\{\}leq \textbackslash\{\}frac\{5\textbackslash\{\}pi\}\{6\} \textbackslash\{\}text\{ and \} x \textbackslash\{\}neq \textbackslash\{\}frac\{\textbackslash\{\}pi\}\{3\}\textbackslash\{\}\}\$ \\

    D\colorbox[HTML]{FFFFFF}{\strut :} \$\textbackslash\{\}\{x\textbar{}-\textbackslash\{\}frac\{5\textbackslash\{\}pi\}\{6\} \textbackslash\{\}leq x \textbackslash\{\}leq \textbackslash\{\}frac\{5\textbackslash\{\}pi\}\{6\} \textbackslash\{\}text\{ and \} x \textbackslash\{\}neq \textbackslash\{\}frac\{\textbackslash\{\}pi\}\{3\}\textbackslash\{\}\}\$\colorbox[HTML]{FFEDED}{\strut <|Assistant|>}\colorbox[HTML]{FFDADA}{\strut <think>} \\
    \vspace{2pt}
    \hrule \vspace{2pt}
    \textbf{Response:} \colorbox[HTML]{FFDDDD}{\strut <think>} \\
    Okay\colorbox[HTML]{FFD9D9}{\strut ,} so I need to solve this logarithmic inequality: log base 1/2 of \textbar{}x - $\pi$\colorbox[HTML]{FFFFFF}{\strut /}3\colorbox[HTML]{FFFAFA}{\strut \textbar{}} is greater than or equal to log base 1/2 of $\pi$/2\colorbox[HTML]{FFE2E2}{\strut .} Then choose the correct solution set from the options A to D\colorbox[HTML]{FFDFDF}{\strut .} Let me think step\colorbox[HTML]{FFFFFF}{\strut  by} step\colorbox[HTML]{FFF3F3}{\strut .} \\

    First, I remember that logarithmic functions have certain properties, especially when dealing with inequalities. Since the base here is 1/2, which is between 0\colorbox[HTML]{FFFDFD}{\strut  and} 1, the logarithmic function is decreasing. That means when I apply it to both sides of an inequality, the direction of the inequality will reverse\colorbox[HTML]{FFEDED}{\strut .} So, if log\_b(a) $\geq$ log\_b(c\colorbox[HTML]{FFF2F2}{\strut ),} then for 0 < b < 1\colorbox[HTML]{FFFBFB}{\strut ,} a\colorbox[HTML]{FFFCFC}{\strut  $\leq$} c\colorbox[HTML]{FFEFEF}{\strut .} Right\colorbox[HTML]{FFEDED}{\strut ?} Let me confirm that. Yes, because as the input to a decreasing function increases\colorbox[HTML]{FFF6F6}{\strut ,}\colorbox[HTML]{FFFFFF}{\strut  the} output decreases. So, the original inequality log\_\{1/2\}(A) $\geq$ log\_\{1/2\colorbox[HTML]{FFFBFB}{\strut \}(}B) implies A $\leq$ B\colorbox[HTML]{FFE9E9}{\strut .} \\

    Therefore, applying this property here, the given inequality log\_\{1/2\colorbox[HTML]{FFF6F6}{\strut \}\textbar{}}x\colorbox[HTML]{FFFAFA}{\strut  -} $\pi$\colorbox[HTML]{FFF9F9}{\strut /}3\colorbox[HTML]{FFFDFD}{\strut \textbar{}} $\geq$ log\_\{1/2\colorbox[HTML]{FFF1F1}{\strut \}(}$\pi$\colorbox[HTML]{FFFCFC}{\strut /2}) implies that \textbar{}x - $\pi$/3\textbar{} $\leq$ $\pi$/\colorbox[HTML]{FFF8F8}{\strut 2}\colorbox[HTML]{FFE5E5}{\strut .} But wait, also, we need to consider the domain of the logarithm. The argument of the logarithm must be positive\colorbox[HTML]{FFFAFA}{\strut .} So, \textbar{}x - $\pi$\colorbox[HTML]{FFFAFA}{\strut /}3\colorbox[HTML]{FFFCFC}{\strut \textbar{}}\colorbox[HTML]{FFFDFD}{\strut  >} 0\colorbox[HTML]{FFF5F5}{\strut .} Which means x\colorbox[HTML]{FFE9E9}{\strut  $\neq$}\colorbox[HTML]{FFFEFE}{\strut  $\pi$}\colorbox[HTML]{FFFBFB}{\strut /}\colorbox[HTML]{FFFDFD}{\strut 3}\colorbox[HTML]{FFCDCD}{\strut .} Because absolute value is always non-negative\colorbox[HTML]{FFECEC}{\strut ,} but logarithm is\colorbox[HTML]{FFFEFE}{\strut  undefined} at 0\colorbox[HTML]{FFEFEF}{\strut .} So x\colorbox[HTML]{FFF1F1}{\strut  cannot}\colorbox[HTML]{FFF5F5}{\strut  be}\colorbox[HTML]{FFEEEE}{\strut  $\pi$}\colorbox[HTML]{FFF6F6}{\strut /}\colorbox[HTML]{FFFCFC}{\strut 3}\colorbox[HTML]{FFE2E2}{\strut .} \\

    So, putting this together, the solution set is the set of all x such that \textbar{}x\colorbox[HTML]{FFF8F8}{\strut  -}\colorbox[HTML]{FFFDFD}{\strut  $\pi$}\colorbox[HTML]{FFF8F8}{\strut /}\colorbox[HTML]{FFFCFC}{\strut 3}\colorbox[HTML]{FFF4F4}{\strut \textbar{}}\colorbox[HTML]{FFF8F8}{\strut  $\leq$} $\pi$\colorbox[HTML]{FFFEFE}{\strut /}\colorbox[HTML]{FFDDDD}{\strut 2}\colorbox[HTML]{FFD8D8}{\strut  and} x\colorbox[HTML]{FFEBEB}{\strut  $\neq$} $\pi$/3\colorbox[HTML]{FFD8D8}{\strut .} Now, let's solve \textbar{}x\colorbox[HTML]{FFFDFD}{\strut  -}\colorbox[HTML]{FFFBFB}{\strut  $\pi$}\colorbox[HTML]{FFF8F8}{\strut /}\colorbox[HTML]{FFFBFB}{\strut 3}\colorbox[HTML]{FFFCFC}{\strut \textbar{}} $\leq$ $\pi$/2\colorbox[HTML]{FFEFEF}{\strut .}\colorbox[HTML]{FFFBFB}{\strut  The} absolute value inequality \textbar{}\colorbox[HTML]{FFF3F3}{\strut A}\colorbox[HTML]{FFFBFB}{\strut \textbar{}}\colorbox[HTML]{FFF7F7}{\strut  $\leq$} B\colorbox[HTML]{FFE2E2}{\strut  (}\colorbox[HTML]{FFF9F9}{\strut where} B\colorbox[HTML]{FFE2E2}{\strut  >}\colorbox[HTML]{FFFCFC}{\strut  }0\colorbox[HTML]{FFFCFC}{\strut )}\colorbox[HTML]{FFF6F6}{\strut  is} equivalent\colorbox[HTML]{FFF9F9}{\strut  to} -B\colorbox[HTML]{FFEDED}{\strut  $\leq$} A\colorbox[HTML]{FFE9E9}{\strut  $\leq$} B\colorbox[HTML]{FFDCDC}{\strut .} So here, -\colorbox[HTML]{FFF8F8}{\strut $\pi$}\colorbox[HTML]{FFF7F7}{\strut /}\colorbox[HTML]{FFF2F2}{\strut 2} $\leq$ x\colorbox[HTML]{FFF8F8}{\strut  -}\colorbox[HTML]{FFFDFD}{\strut  $\pi$}\colorbox[HTML]{FFF6F6}{\strut /}\colorbox[HTML]{FFE5E5}{\strut 3}\colorbox[HTML]{FFF1F1}{\strut  $\leq$} $\pi$/\colorbox[HTML]{FFF7F7}{\strut 2}\colorbox[HTML]{FFDCDC}{\strut .} Let\colorbox[HTML]{FFF9F9}{\strut 's} solve for x\colorbox[HTML]{FFE5E5}{\strut .} \\

    First\colorbox[HTML]{FFFEFE}{\strut ,} add $\pi$/3 to all parts\colorbox[HTML]{FFF9F9}{\strut :} \\

    -$\pi$/2\colorbox[HTML]{FFF8F8}{\strut  +} $\pi$/3\colorbox[HTML]{FFFFFF}{\strut  $\leq$} x\colorbox[HTML]{FFF8F8}{\strut  $\leq$} $\pi$/2\colorbox[HTML]{FFF1F1}{\strut  +} $\pi$/\colorbox[HTML]{FFFCFC}{\strut 3} \\

    Compute -$\pi$/2 + $\pi$/3\colorbox[HTML]{FFE0E0}{\strut .} To add these, find a common denominator\colorbox[HTML]{FFECEC}{\strut .} 6 is\colorbox[HTML]{FFFCFC}{\strut  the} least common multiple of 2\colorbox[HTML]{FFF6F6}{\strut  and} 3\colorbox[HTML]{FFEDED}{\strut .} So\colorbox[HTML]{FFFCFC}{\strut :} \\

    -$\pi$/2\colorbox[HTML]{FFECEC}{\strut  =} -3\colorbox[HTML]{FFFDFD}{\strut $\pi$}/\colorbox[HTML]{FFF4F4}{\strut 6} \\

    $\pi$/3\colorbox[HTML]{FFFAFA}{\strut  =} 2\colorbox[HTML]{FFFFFF}{\strut $\pi$}/\colorbox[HTML]{FFFBFB}{\strut 6} \\

    So\colorbox[HTML]{FFFDFD}{\strut ,} -3$\pi$/6 + 2$\pi$/6 = (-3\colorbox[HTML]{FFFFFF}{\strut $\pi$}\colorbox[HTML]{FFFBFB}{\strut  +} 2$\pi$)/6\colorbox[HTML]{FFFAFA}{\strut  =} -$\pi$/6 \\

    Similarly\colorbox[HTML]{FFFFFF}{\strut ,} $\pi$/2 + $\pi$\colorbox[HTML]{FFFEFE}{\strut /}3\colorbox[HTML]{FFE2E2}{\strut :} \\

    $\pi$/2\colorbox[HTML]{FFFAFA}{\strut  =} 3$\pi$/\colorbox[HTML]{FFFDFD}{\strut 6} \\

    $\pi$/3\colorbox[HTML]{FFFBFB}{\strut  =}\colorbox[HTML]{FFFEFE}{\strut  }2$\pi$/6 \\

    3$\pi$/6 + 2$\pi$/6 = 5$\pi$/\colorbox[HTML]{FFFAFA}{\strut 6} \\

    Therefore, the inequality \textbar{}x - $\pi$/3\textbar{} $\leq$ $\pi$/2 simpl\colorbox[HTML]{FFFFFF}{\strut ifies} to -$\pi$/6 $\leq$ x\colorbox[HTML]{FFFEFE}{\strut  $\leq$} 5$\pi$/6\colorbox[HTML]{FFE8E8}{\strut .} But remember x\colorbox[HTML]{FFEBEB}{\strut  $\neq$} $\pi$/3\colorbox[HTML]{FFFEFE}{\strut  because} \textbar{}x - $\pi$/3\textbar{} > 0\colorbox[HTML]{FFDBDB}{\strut .} \\

    So, the solution set is all x in\colorbox[HTML]{FFF2F2}{\strut  [-}$\pi$/6\colorbox[HTML]{FFEEEE}{\strut ,} 5$\pi$/\colorbox[HTML]{FFF4F4}{\strut 6}\colorbox[HTML]{FFEAEA}{\strut ]}\colorbox[HTML]{FFEDED}{\strut  except} x\colorbox[HTML]{FFF0F0}{\strut  =} $\pi$/3\colorbox[HTML]{FFE5E5}{\strut .} Now, looking at the options: \\

    A: \{x \textbar{} -$\pi$/6\colorbox[HTML]{FFFEFE}{\strut  $\leq$} x $\leq$ 5$\pi$/6\colorbox[HTML]{FFE3E3}{\strut \}}\colorbox[HTML]{FFE0E0}{\strut  $\to$} Doesn't\colorbox[HTML]{FFEDED}{\strut  exclude} $\pi$/3\colorbox[HTML]{FFDEDE}{\strut ,} so incorrect\colorbox[HTML]{FFFAFA}{\strut .} \\

    B: \{x\colorbox[HTML]{FFFCFC}{\strut  \textbar{}} x\colorbox[HTML]{FFFEFE}{\strut  $\leq$} -$\pi$/6 or x $\geq$ 5$\pi$/6\}\colorbox[HTML]{FFFAFA}{\strut  $\to$} This is the complement of\colorbox[HTML]{FFFBFB}{\strut  the}\colorbox[HTML]{FFFFFF}{\strut  interval}\colorbox[HTML]{FFF1F1}{\strut ,} which would be if\colorbox[HTML]{FFFEFE}{\strut  the} inequality was reversed\colorbox[HTML]{FFF0F0}{\strut .} But since our inequality was \textbar{}x - $\pi$/3\textbar{} $\leq$ $\pi$/2\colorbox[HTML]{FFF2F2}{\strut ,} this is the outside\colorbox[HTML]{FFEDED}{\strut ,} which is wrong\colorbox[HTML]{FFF2F2}{\strut .} \\

    C: \{x\colorbox[HTML]{FFF7F7}{\strut  \textbar{}} -$\pi$/6\colorbox[HTML]{FFF7F7}{\strut  $\leq$} x $\leq$ 5$\pi$/6\colorbox[HTML]{FFDDDD}{\strut  and} x\colorbox[HTML]{FFE7E7}{\strut  $\neq$} $\pi$/\colorbox[HTML]{FFFEFE}{\strut 3}\colorbox[HTML]{FFF6F6}{\strut \}}\colorbox[HTML]{FFE5E5}{\strut  $\to$} This matches our solution\colorbox[HTML]{FFDFDF}{\strut .}\colorbox[HTML]{FFFBFB}{\strut  Includes}\colorbox[HTML]{FFE8E8}{\strut  the} interval\colorbox[HTML]{FFF6F6}{\strut  and}\colorbox[HTML]{FFF5F5}{\strut  excludes} $\pi$/3\colorbox[HTML]{FFF0F0}{\strut .} \\

    D: \{x\colorbox[HTML]{FFFAFA}{\strut  \textbar{}} -5$\pi$\colorbox[HTML]{FFFEFE}{\strut /}6\colorbox[HTML]{FFEEEE}{\strut  $\leq$} x $\leq$ 5$\pi$/6\colorbox[HTML]{FFEBEB}{\strut  and} x\colorbox[HTML]{FFECEC}{\strut  $\neq$}\colorbox[HTML]{FFFCFC}{\strut  $\pi$}/\colorbox[HTML]{FFFFFF}{\strut 3}\colorbox[HTML]{FFF2F2}{\strut \}}\colorbox[HTML]{FFD9D9}{\strut  $\to$}\colorbox[HTML]{FFF3F3}{\strut  The} interval\colorbox[HTML]{FFFBFB}{\strut  here}\colorbox[HTML]{FFEAEA}{\strut  is} from -5$\pi$/6 to 5$\pi$/6\colorbox[HTML]{FFE1E1}{\strut ,} which is different\colorbox[HTML]{FFE3E3}{\strut .} Let me check if I made a mistake\colorbox[HTML]{FFFFFF}{\strut  in} solving\colorbox[HTML]{FFFCFC}{\strut  the} inequality\colorbox[HTML]{FFE0E0}{\strut .} \\

    \colorbox[HTML]{FFF5F5}{\strut Wait}\colorbox[HTML]{FFEBEB}{\strut ,} when solving \textbar{}x - $\pi$/3\textbar{} $\leq$ $\pi$/2\colorbox[HTML]{FFF8F8}{\strut ,} adding $\pi$/3 to -$\pi$/2 gives -$\pi$/6\colorbox[HTML]{FFD9D9}{\strut ,}\colorbox[HTML]{FFFEFE}{\strut  and} adding $\pi$/3 to $\pi$/2 gives 5$\pi$/6\colorbox[HTML]{FFD5D5}{\strut .} That\colorbox[HTML]{FFFEFE}{\strut 's} correct\colorbox[HTML]{FFDEDE}{\strut .} So D\colorbox[HTML]{FFEEEE}{\strut  is} incorrect\colorbox[HTML]{FFFFFF}{\strut  because}\colorbox[HTML]{FFE3E3}{\strut  the} interval\colorbox[HTML]{FFE4E4}{\strut  is} from -5$\pi$/6\colorbox[HTML]{FFC8C8}{\strut ,} which\colorbox[HTML]{FFF3F3}{\strut  is} more\colorbox[HTML]{FFFCFC}{\strut  negative}\colorbox[HTML]{FFD7D7}{\strut .} So that's\colorbox[HTML]{FFFCFC}{\strut  not} matching\colorbox[HTML]{FFEEEE}{\strut .} So\colorbox[HTML]{FFF1F1}{\strut  the} answer must\colorbox[HTML]{FFFAFA}{\strut  be} C\colorbox[HTML]{FFD7D7}{\strut .} \\

    \colorbox[HTML]{FFFAFA}{\strut But} let\colorbox[HTML]{FFF6F6}{\strut  me} just\colorbox[HTML]{FFFDFD}{\strut  verify} once\colorbox[HTML]{FFFFFF}{\strut  more}\colorbox[HTML]{FFE0E0}{\strut .}\colorbox[HTML]{FFEBEB}{\strut  The} original inequality\colorbox[HTML]{FFEEEE}{\strut  is}\colorbox[HTML]{FFF9F9}{\strut  log}\colorbox[HTML]{FFAAAA}{\strut \_\{}\colorbox[HTML]{FFF4F4}{\strut 1}\colorbox[HTML]{FFD900}{\strut /}
    \par}
\end{tcolorbox}

\section{Inference-Time Segment State Update}
\label{app:segment}

We follow the segment-level state correction of~\citet{zhang2026statetransitionframeworkefficientllm} (Liang~Zhang et~al.) and restate it briefly for completeness. During inference, noisy reasoning segments can mislead subsequent generation and push LA state transitions off a stable trajectory. The correction uses a global reasoning direction with a constant coefficient $\alpha$.

We partition the generated token sequence into segments of fixed size $k$. Let $\mathbf{S}^{(l)}_{\mathrm{tok}}$ denote the LA state after $\mathrm{tok}$ generated tokens in layer $l$. The state change (reasoning direction) of the $t$-th segment is
\begin{equation}
    \nabla_t^l = \mathbf{S}^{(l)}_{k \cdot t} - \mathbf{S}^{(l)}_{k \cdot (t-1)}.
\end{equation}
Noisy steps typically produce $\nabla_t$ that deviate from the global trend. At the end of segment $t$, we first aggregate prior \emph{corrected} directions via a running average,
\begin{equation}
    \bar{\nabla}_{t-1}^l = \frac{1}{t-1} \sum_{i=1}^{t-1} \hat{\nabla}_i^l,
\end{equation}
with $\bar{\nabla}_0^l$ a zero matrix, and then mix the current raw direction with this history,
\begin{equation}
    \hat{\nabla}_t^l = (1 - \alpha) \nabla_t^l + \alpha \bar{\nabla}_{t-1}^l,
\end{equation}
where $\alpha$ is a fixed correction coefficient. Finally, the segment-level state used going forward is
\begin{equation}
    \mathbf{S}^{(l)}_{k\cdot t} \leftarrow \mathbf{S}^{(l)}_{k\cdot (t-1)} + \hat{\nabla}_t^l.
\end{equation}
This pulls the current segment update toward historical trends and limits accumulation of local noise. We follow~\citet{zhang2026statetransitionframeworkefficientllm} for this correction rule; LISA's novelty is only in applying it jointly with the Indexer-guided sparse branch.

\end{document}